\documentclass[conference]{IEEEtran}
\IEEEoverridecommandlockouts
\usepackage{cite}
\usepackage{amsmath,amssymb,amsfonts}
\usepackage{algorithmic}
\usepackage{verbatim}
\usepackage{graphicx}
\usepackage{textcomp}
\usepackage{xcolor}
\usepackage{multirow}
\usepackage{caption}
\usepackage{subcaption}
\def\BibTeX{{\rm B\kern-.05em{\sc i\kern-.025em b}\kern-.08em
    T\kern-.1667em\lower.7ex\hbox{E}\kern-.125emX}}
\usepackage{balance}
\usepackage{url}
\usepackage{subcaption}
\usepackage{booktabs}
\newtheorem{definition}{Definition}

\usepackage[lined,boxed,vlined,ruled,linesnumbered]{algorithm2e}

\usepackage{listings}
\usepackage{amssymb}
\usepackage{multirow}
\usepackage{enumitem}
\usepackage{tikz}

\newcommand{\alglinenonumber}[1]{\item[] #1}

\def\BibTeX{{\rm B\kern-.05em{\sc i\kern-.025em b}\kern-.08em
    T\kern-.1667em\lower.7ex\hbox{E}\kern-.125emX}}
\begin{document}

\title{GraphRARE: Reinforcement Learning Enhanced Graph Neural Network with Relative Entropy}

\author{\IEEEauthorblockN{Tianhao Peng\IEEEauthorrefmark{2}, Wenjun Wu\IEEEauthorrefmark{2}\IEEEauthorrefmark{1}, Haitao Yuan\IEEEauthorrefmark{3}\IEEEauthorrefmark{1}, Zhifeng Bao\IEEEauthorrefmark{4}, Zhao Pengrui\IEEEauthorrefmark{2}, Xin Yu\IEEEauthorrefmark{2}, \\Xuetao Lin\IEEEauthorrefmark{2}, Yu Liang\IEEEauthorrefmark{5}, Yanjun Pu\IEEEauthorrefmark{2}}
\IEEEauthorblockA{\IEEEauthorrefmark{2}Beihang University, China \IEEEauthorrefmark{3}Nanyang Technological University, Singapore \\\IEEEauthorrefmark{4}RMIT University, Australia \IEEEauthorrefmark{5}Beijing University of Technology, China\\
\IEEEauthorrefmark{2}\{pengtianhao,wwj09315,zhaopengrui,nlsdeyuxin,xtlin,buaapyj\}@buaa.edu.cn}\IEEEauthorrefmark{3}haitao.yuan@ntu.edu.sg,\IEEEauthorrefmark{4}zhifeng.bao@rmit.edu.au,\IEEEauthorrefmark{5}yuliang@bjut.edu.cn
\thanks{\IEEEauthorrefmark{1}Both Wenjun Wu and Haitao Yuan are the corresponding authors.}
}
\maketitle

\begin{tikzpicture}[remember picture,overlay]
\node[anchor=north,yshift=-1cm] at (current page.north) {2024 IEEE 40th International Conference on Data Engineering (ICDE)};
\end{tikzpicture}

\begin{abstract}

Graph neural networks (GNNs) have shown advantages in graph-based analysis tasks. However, most existing methods have the homogeneity assumption and show poor performance on heterophilic graphs, where the linked nodes have dissimilar features and different class labels, and the semantically related nodes might be multi-hop away. To address this limitation, this paper presents GraphRARE, a general framework built upon node relative entropy and deep reinforcement learning, to strengthen the expressive capability of GNNs. An innovative node relative entropy, which considers node features and structural similarity, is used to measure mutual information between node pairs. In addition, to avoid the sub-optimal solutions caused by mixing useful information and noises of remote nodes, a deep reinforcement learning-based algorithm is developed to optimize the graph topology. This algorithm selects informative nodes and discards noisy nodes based on the defined node relative entropy. Extensive experiments are conducted on seven real-world datasets. The experimental results demonstrate the superiority of GraphRARE in node classification and its capability to optimize the original graph topology.
\end{abstract}


\section{Introduction}\label{sec:intro}

Graph data structures have been widely used in many real-world scenarios such as database management system~\cite{DBLP:journals/pvldb/ShenHZ23, DBLP:conf/icde/LiTL23, DBLP:conf/vldb/Horchidan23}, knowledge graphs~\cite{DBLP:conf/icde/Zheng0QYCZ23, DBLP:conf/icde/ZhangWYZCZ23}, recommendation systems~\cite{zhou2020graph, DBLP:conf/icde/WuWXFGWSZWZ23, DBLP:conf/kdd/0009ZGZNQH19, li2023mhrr}, and traffic forecasting~\cite{DBLP:conf/icde/00020BF21, DBLP:journals/dase/YuanL21, DBLP:journals/pvldb/00020B22}. In recent years, graph neural networks (GNNs) have exhibited advantages in numerous graph-based analytic tasks, including node classification, edge prediction, graph classification, and graph clustering~\cite{zheng2022graph}. 

Most GNNs are based on a message-passing neural network (MPNN) framework, which aggregates features of neighboring nodes. Such a design of node message aggregation is effective under the assumption of homophily, which requires that adjacent nodes contain similar features or belong to the same class labels~\cite{zheng2022graph}. 
Meanwhile, heterophilic graphs have also been widespread in the real world~\cite{DBLP:conf/icde/ZhuZWXY023, DBLP:conf/icde/GuZBCZY23, DBLP:journals/corr/abs-2311-09261, li2023ltrgcn, peng2023clgt, DBLP:conf/www/LiYFMXW23, DBLP:conf/icde/Han00S21}. In these graphs, linked nodes have different features and class labels, whereas semantically related nodes can be multi-hop away. For instance, different types of amino acids are more likely to connect together in protein structures, fraudsters are more likely to build connections with customers instead of other fraudsters in online purchasing networks. Therefore, local neighbors of the original graph topology fail to capture informative nodes at a long distance, which may introduce noises, resulting in poor MPNN performance~\cite{DBLP:conf/aaai/HeLLWJ022}. Thus, capturing and extracting important features from distant but informative nodes can enhance the performance of MPNNs~\cite{abu2019mixhop, DBLP:conf/nips/ZhuYZHAK20, DBLP:conf/nips/JinYHWWHH21, zheng2022graph, DBLP:conf/icde/ZhaoYT21, DBLP:conf/kdd/ZhangZY0023}.

\begin{figure}[!t]
	\centering 
	\includegraphics[width=0.8\linewidth]{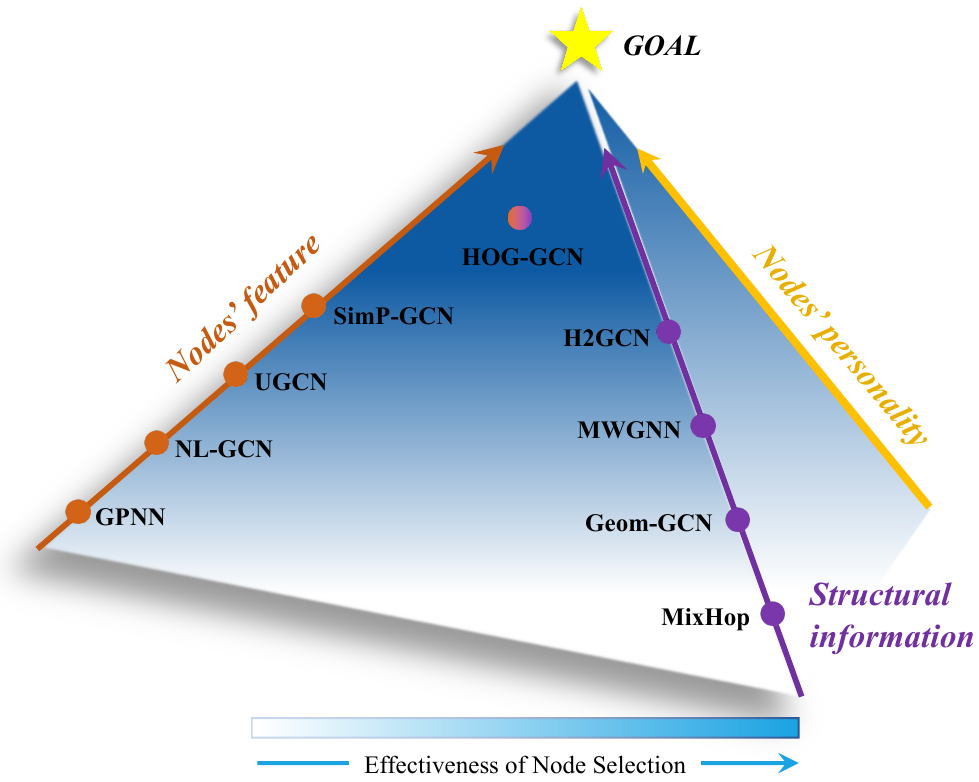} 
	\caption{The visual illustration of related works on heterophilic graph neural networks. The color shades indicate the effectiveness of node selection in reconstructing the original graph topology, with darker shades representing higher effectiveness.}
	\label{fig:intro}
\end{figure}

In recent years, extensive efforts have been made to exploit remote nodes' information under the heterophily setting. 
The main idea is to optimize the graph topology based on the homophily assumption that message aggregation mechanism works the  best in homophilic graphs~\cite{DBLP:conf/www/FangXSLZ22, DBLP:conf/icdm/YanHSYK22, DBLP:conf/wsdm/JinDW0LT21, yang2022graph, DBLP:conf/iclr/PeiWCLY20, DBLP:conf/www/DuSFMLHZ22, feng2023towards, DBLP:journals/tkde/YangGLZCW23}.
It is widely known that there are two steps to optimize the original topology and reconstruct graphs. The first step is to select an appropriate metric to measure the remote nodes' importance as accurately as possible. The second step is to connect the ego node and a small number of remote nodes based on the importance degree. Existing studies mainly leverage inherent graph features, such as node features and structure similarity, to select relative nodes with a constant number, and then focus on the optimization of the first step. Unfortunately, the graph reconstruction in the second step is often neglected. As Figure~\ref{fig:intro} shows, the goal of GNN in heterophilic graphs is to make full use of both the node feature and structural feature of the graph, and consider the personality of each node to improve the accuracy of selecting important nodes. UGCN~\cite{DBLP:conf/nips/JinYHWWHH21}, NL-GNN~\cite{DBLP:journals/corr/abs-2005-14612}, SimP-GCN~\cite{DBLP:conf/wsdm/JinDW0LT21} and GPNN~\cite{yang2022graph} mainly utilize feature similarity between nodes as a metric to reconstruct the neighbor set. 
{GBK-GNN~\cite{DBLP:conf/www/DuSFMLHZ22} utilizes kernels to measure the feature similarity between node pairs to guide the aggregation operation.}
On the contrary, other studies consider the topology of graphs. For example, MixHop~\cite{abu2019mixhop} and H2GCN~\cite{DBLP:conf/nips/ZhuYZHAK20} utilize the multi-hop node information in each convolution layer. Geom-GCN~\cite{DBLP:conf/iclr/PeiWCLY20} maps the original graph to a latent space and defines the geometric relationship as a criterion to reconstruct the original graph topology. MWGNN~\cite{DBLP:conf/www/MaCRSW22} considers the structural similarity to select remote nodes. However, the aforementioned methods rarely consider node features and structural information at the same time, and hence fail to make full use of the rich information of graphs. 
Hence, Polar-GNN~\cite{DBLP:conf/www/FangXSLZ22} and HOG-GCN~\cite{DBLP:conf/aaai/WangJWHH22} consider both node features and structure to guide the aggregation operation of GNNs. Polar-GNN attempts to model pairwise similarities and dissimilarities of nodes, but it fails to capture higher-order structural information or make full use of it. HOG-GCN uses the label propagation technique to estimate the homophily degree matrix, which is affected by the number of labeled nodes, hence making it difficult to fully utilize the structural information. 
Regarding the optimization for the original graph topology, existing studies do not design appropriate methods for optimization but simply set to hyperparameters. Examples like UGCN~\cite{DBLP:conf/nips/JinYHWWHH21}, SimP-GCN~\cite{DBLP:conf/wsdm/JinDW0LT21}, and MI-GCN~\cite{conf/www/migcn} select the $top\text{-}k$ most similar remote nodes to construct new neighboring node sets or remove the $top\text{-}d$ most dissimilar ones from neighboring sets. These methods rely on prior knowledge or require a huge amount of human efforts to tune the hyper-parameters, highlighting the demand for improved methods in the current approach.

In summary, the aforementioned methods cannot fill the gap in the field of heterophilic graphs, and the major challenges can be summarized as follows: 

\begin{itemize}[leftmargin=*]
\item {\textit{Appropriate Metric}. The key problem in extending higher-hop neighbors is to determine an appropriate metric for measuring the node importance when extending higher-hop neighbors. Most existing studies leverage inherent structural information directly from the original graphs to measure node importance. However, \textbf{how to integrate both node features and structural information} to reconstruct the original graph topology and enhance the expressive power of GNNs remains an ongoing challenge.}
\item \textit{Personality of nodes}. During the process of reconstructing an enhanced graph from the original graph, it is often necessary to set the hyper-parameters, $k$ and $d$, where we need to select top-$k$ important nodes from the higher-order neighbors as the first-order neighbors or delete top-$d$ nodes from the first-order neighbors for each node. However, \textbf{the values of $k$ and $d$ ought to vary across individual nodes}, especially in the case of heterogeneous graphs. Therefore, it becomes another challenge to consider the individual characteristics of nodes and set different values of the hyper-parameters.
\end{itemize}

To address the above challenges, we propose a novel approach, \textbf{GraphRARE} (\textbf{R}einforcement le\textbf{A}rning enhanced Graph Neural Network with \textbf{R}elative \textbf{E}ntropy), specifically designed for heterophilic graphs. 
GraphRARE aims to adaptively optimize the graph topology while leveraging the advantages of advanced Graph Neural Networks (GNNs), which is critical in the field of database systems with inherently graph-like structures~\cite{DBLP:journals/pvldb/ShenHZ23, DBLP:conf/icde/LiTL23, DBLP:conf/vldb/Horchidan23, DBLP:conf/icde/Zheng0QYCZ23}.

At first, we define a node-aware relative entropy to measure node importance. In particular, this entropy is composed of node feature entropy and node structural entropy, which are designed to capture node features and structural information, respectively. Therefore, the proposed relative entropy is an appropriate metric, \textbf{addressing the first challenge}. Next, for each node, we calculate the node relative entropy between the ego node and its remote neighbors. Subsequently, we construct a descending-order node sequence based on the importance of remote neighbor nodes. \textbf{To address the second challenge}, we consider the personality of nodes and employ a Deep Reinforcement Learning (DRL)-based algorithm to determine the values of hyper-parameters $k$ and $d$ for each node. At last, we jointly train the DRL-based algorithm and an existing advanced GNN model, enabling the optimization of the graph topology and facilitating the downstream task (i.e., node classification in this paper). 

The main technical contributions of this study can be summarized as follows:

\begin{itemize}
\item {A node relative entropy is defined to measure the similarity between nodes based on their structure and features, which enhances the application of node entropy theory in the domain of graph data (Sec.~\ref{sec:rec}).}
\item {The node personality is captured when enhancing the graph topology, where we use a well-designed reinforcement learning model to set  different hyper-parameter values for different nodes (Sec.~\ref{sec:drl}).}
\item {The GraphRARE framework is developed for heterophilic graphs based on the proposed node relative entropy and deep reinforcement learning (Sec.~\ref{sec:framework}). The combination of DRL and GNNs provides end-to-end training to optimize the original graph topology (Sec.~\ref{sec:nfl}).}
\item {Extensive experiments have been conducted and the results show that the proposed method represents a general yet useful framework, which can enhance the GNN performance in node classification (Sec.~\ref{sec:exp}).} 
\end{itemize}

\begin{figure*}[!t]
	\centering 
	\includegraphics[width=\linewidth]{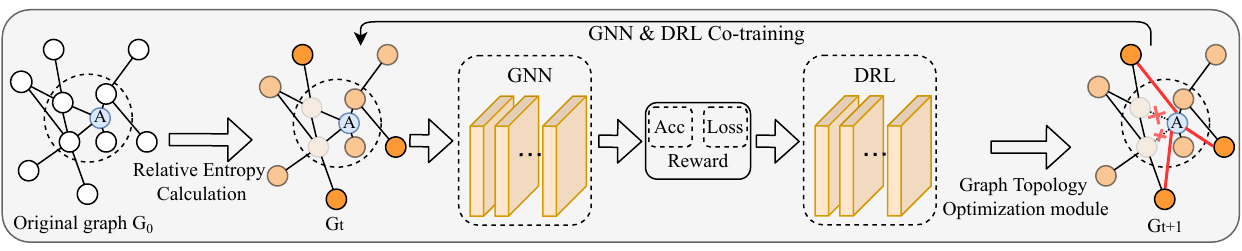} 
	\caption{An illustration of the proposed GraphRARE framework. The GNN and DRL module are trained jointly to optimize the graph topology based on the relative entropy (the depth of the node's color indicates the value of entropy).} 
	\label{fig:framework}
\end{figure*}

\section{Preliminary}

\begin{table}[!t]
    \caption{Symbols and Definitions.}
    \renewcommand\arraystretch{1.05}
    \centering
    \small
    \scalebox{0.9}{
    \begin{tabular}{ll}
    \toprule
    \textbf{Symbol} & \textbf{Definition}\\
    \midrule
    $G\text{=}(V, E, X, A)$ & Graphs\\
    $N$ & Number of nodes in the graph\\
    $V\text{=}\{ v_1, v_2, ..., v_N\}$ & Set of nodes\\
    $E \subseteq V \times V$ & Set of edges\\
    $X \in \mathbb{R}^{N \times d}$ & Node feature matrix\\
    $A \in \mathbb{R}^{N \times N}$ & Adjacency matrix\\
    $\mathcal{H}$ & Homophily ratio of a graph\\
    $H(v,u)$ & Node relative entropy between node $v$ and $u$\\
    $H_f(v,u)$  & Feature Entropy between node $v$ and $u$\\
    $H_s(v,u)$ & Structural entropy between node $v$ and $u$\\
    $\lambda$ & The hyper-parameter in node relative entropy\\
    $S_t$ & Multi-discrete state at step $t$\\
    $A_t$ & Multi-discrete action at step $t$\\
    $R(S_t)$ & Reward of state $S_t$\\
    $y_v$ & Class of node $v$\\
    $\mathcal N_k(v)$ & The $k$-hop neighbors of node $v$\\
    $\mathcal N_1(v)$ & The first-order neighbors of node $v$\\
    \bottomrule
    \end{tabular}
    }
    \label{table:symbols}
\end{table}

In this section, we first introduce some basic concepts and then formulate the problem of heterophilic graph topology optimization. In addition, frequently used symbols are summarized in Table~\ref{table:symbols}.

\subsection{Basic Concepts}
\noindent \textbf{Graph.} A graph can be formulated as $G\text{=}(V, E, X, A)$, where $V\text{=}\{ v_1, v_2, ..., v_N\}$ denotes a node set, $E \subseteq V \times V$ is an edge set, $X \in \mathbb{R}^{N \times d}$ represents node features, and $A \in \mathbb{R}^{N \times N}$ is the adjacency matrix; $N$ is the number of nodes, and $d$ is the dimension of a node feature. Particularly, $a_{ij}\text{=}1$ indicates that an edge between nodes $v_i$ and $v_j$ exists; otherwise, $a_{ij}\text{=}0$. Further, $x_i$ represents the feature vector of a node $v_i$. The degree of a node $v$ is denoted by $d_v$. The $k$-hop neighbors of a node $v$ are denoted by $\mathcal N_k(v)$. For instance, a directly connected (one-hop) neighbor set of a node $v$ is $\mathcal N_1(v)$. 

\noindent \textbf{Heterophilic Graph and Homophily Ratio.} Different nodes in a graph may correspond to different categories/classes, so we call it a heterophilic graph, where the linked nodes belong to different categories/classes or possess dissimilar features. In addition, the definition of heterophilic graphs can be measured by the homophily ratio, which is computed based on its edge homophily~\cite{DBLP:conf/nips/ZhuYZHAK20}:
\begin{flalign} 
\mathcal{H} &= \frac{|\{(v, u) \in {E}: y_{v}=y_{u}\}|}{|{E}|}
\end{flalign}
Here, $\mathcal{H}$ is the homophily ratio, $y_v$ is the label of node $v$. The value range of $\mathcal{H}$ is $[0,1]$. A large $\mathcal{H}$ indicates a homophilic graph, whereas a low $\mathcal{H}$ indicates a heterophilic graph. 

\noindent \textbf{Topology Optimization for Heterophilic Graph Learning.} The majority of Graph Neural Networks (GNNs) follow the homophily assumption, which aggregates messages extracted from local neighbor nodes, but this approach may not effectively generalize to heterophilic graphs. Improving the performance of traditional GNNs (e.g., GCN, GAT, GraphSAGE) in such graphs is a challenging task. One general and effective method is to optimize the original graph topology (i.e., adding edges to link nodes or deleting existing links) using a well-designed metric to increase the homophily ratio. Numerous approaches, such as UGCN~\cite{DBLP:conf/nips/JinYHWWHH21}, NL-GNN~\cite{DBLP:journals/corr/abs-2005-14612}, SimP-GCN~\cite{DBLP:conf/wsdm/JinDW0LT21}, and GPNN~\cite{yang2022graph}, have been proposed for this purpose. By optimizing the heterophilic graph topology, the linked nodes in the graphs tend to have more similar features, making them more suitable for traditional GNNs. This enhancement allows the vanilla GNNs to perform better on heterophilic graphs and improve their overall performance and generalization capability.

\noindent 

\subsection{Problem Formulation}
The main goal of graph topology optimization for graph learning is to restructure the graph in a way that promotes information exchange and interactions between nodes of different types. To achieve this, a well-designed metric is used to measure the relevance between node pairs. Noisy edges are deleted, and informative node pairs are connected to enhance the flow of information in the graph. Formally, we can define it as follows.
\begin{definition}[Topology Optimization for GNNs]
Given a heterophilic graph $G=\langle V, E, X, A\rangle$, a graph neural network model (GNN) designed for a downstream task (i.e., node classification in this paper), the objective is to devise an effective transformation approach for creating a refined graph $G'=\langle V, E', X, A'\rangle$. This approach involves the addition or deletion of edges to optimize the performance of the GNN model on the given downstream task.
\end{definition}

\section{Framework\label{sec:framework}}
As shown in Figure~\ref{fig:framework}, we first introduce a highly effective relative entropy function designed to quantify the significance of one node with respect to another. In particular, to capture both nodes' feature and graph structural information, our proposed node relative entropy is comprised of both node feature entropy and node structural entropy. For example, given a node `A' in the original graph, we leverage the entropy value to measure other nodes' relative significance. Hence, we can add/delete edges linking `A' and some nodes with high/low entropy values. However, each node exhibits its distinct characteristics or "personality," and optimizing the graph topology with a uniform addition or deletion of links may not fully consider these individual traits. Consequently, this approach can lead to sub-optimal results in the optimized graph. 

To address this, we propose a novel approach that models the graph topology optimization as a finite horizon Markov Decision Process with multi-discrete states. This allows us to effectively account for the personality of different nodes within the graph. The entire model comprises two neural network modules: the \textit{Graph Neural Network (GNN)}, which handles the downstream task of node classification, and the \textit{Deep Reinforcement Learning (DRL)} module, responsible for determining the number of added or deleted edges for each node. Moreover, we adopt a co-training approach in a loop mode, where the GNN's training loss or accuracy serves as the reward to train the DRL module. Concurrently, the DRL module generates a reconstructed graph, which then becomes the input to the GNN. The entire pipeline operates as follows: prior to model training, we compute the relative entropy between node pairs. During each training step, the GNN's performance is evaluated on the optimized graph at step $t$ ($G_t$). Based on this performance, including accuracy and loss on the training set, the policy network of the DRL is updated. Subsequently, the graph topology optimization module generates the optimized graph $G_{t+1}$ and feeds it into the GNN for the next iteration of co-training.

In summary, by incorporating reinforcement learning into the graph topology optimization process, we aim to better capture and utilize the diverse traits and relationships between nodes, leading to improved performance in various tasks. This framework allows us to tailor the graph topology to suit the specific characteristics of the nodes, enhancing the overall effectiveness of our proposed model.

\section{Methodology}

This section presents our GraphRARE framework for node classification in heterophilic graphs, addressing the main challenges related to metric and personality of nodes, as described in Section~\ref{sec:intro}. The framework, depicted in Figure~\ref{fig:framework}, follows a multi-step approach. Initially, GraphRARE computes the relative entropy between nodes, and subsequently trains the GNN and DRL modules jointly in an end-to-end manner to accomplish the node classification task. In particular, we first define the fundamental concepts of node relative entropy, encompassing both node feature entropy and node structural entropy. Subsequently, we elucidate the construction of a node sequence based on the relative entropy theory. In addition, we develop a graph reconstruction mechanism to optimize the original graph topology based on deep reinforcement learning. Finally, we take advantage of the GNNs in node feature learning to complete the node classification task. 

\subsection{Node Relative Entropy Calculation\label{sec:rec}}
Entropy is an effective method to measure uncertain information in a graph. Relative entropy (Kullback–Leibler divergence)~\cite{kullback1951information} is one measure to quantify differences in probability distributions, which is defined as follows: 

\begin{flalign} 
\label{equation0} D_{KL}(P\Vert Q) = \sum_{i=1}^{n} P(i)\log\frac{P(i)}{Q(i)}
\end{flalign}
where $P_i$ and $Q_i$ are two probability distributions of the event $i$; $n$ is the number of events; $P$ and $Q$ have the same number of components. Notably, the smaller the entropy is, the smaller the difference between two distributions will be. 

However, it is non-trivial to measure the graphs using the original definition of relative entropy where the rich node features and link structures are difficult to quantify. Inspired by~\cite{luo2021graph, zhang2018measure}, our work optimizes the original graph entropy and introduces a node relative entropy that consists of feature entropy and structural entropy to address this issue. To measure the feature similarity between nodes, the graph feature entropy in~\cite{luo2021graph} is optimized in this study by calculating the entropy between any pair of nodes, and the computational complexity is also reduced. Since the feature entropy does not take structural information into account, we introduce structural entropy to measure the structural similarity between nodes. However, the structural entropy in~\cite{zhang2018measure} has difficulty in accurately measuring the semantically related node pairs, and we optimize it by introducing Jensen–Shannon (JS) divergence~\cite{lin1991divergence}. 

\subsubsection{Node Feature Entropy}
Feature entropy measures the difference between two nodes based on their feature vectors. Inspired by the MinGE model~\cite{luo2021graph}, which defines the graph feature entropy, this paper designs a feature entropy for each pair of nodes in graph data. Compared to the graph feature entropy in~\cite{luo2021graph}, the node feature entropy requires lower computation complexity and is able to compute entropy at the node level. Based on the assumption of node embedding, considering that nodes in the same class are more similar, the node embedding dot product of node pairs is used as a node feature entropy, which can be expressed as follows:

\begin{flalign} 
\label{equation3} z_v &= \phi(x_v)\in \mathbb{R}^h, {\forall}v\in V\\
H_f(v,u) &= -P(z_v,z_u)\log P(z_v,z_u) \nonumber\\
\label{equation5} &= -\frac{e^{\left \langle z_v,z_u\right \rangle}}{\sum_{i,j}e^{\left \langle z_i,z_j\right \rangle}}\log \frac{e^{\left \langle z_v,z_u\right \rangle}}{\sum_{i,j}e^{\left \langle z_i,z_j\right \rangle}}
\end{flalign}
where $\phi(\cdot)$ is an embedding function (e.g., multilayer perceptron), which converts original features to vectors; $x_v$ is a node feature of a node $v$; $h$ is the dimension of the node feature embedding $z_v$; $\left \langle \cdot,\cdot \right \rangle$ is the dot product operation; $\sum_{i,j}e^{\left \langle z_i,z_j\right \rangle}$ represents the dot product sum of a pair of nodes in a graph $G$; $H_f(v,u)$ is the node feature entropy between nodes $v$ and $u$. As shown in Eq.~(\ref{equation5}), the larger the node feature entropy is, the more similar the node features are.

\subsubsection{Node Structural Entropy}
Structural entropy measures the difference between two nodes based on their local structure~\cite{zhang2018measure}. Node degree has been one of the most important structural features in the graph structure entropy~\cite{DBLP:journals/entropy/OmarP20}. Therefore, it is reasonable to calculate the node structural entropy by comparing the ordered degree sequences of nodes and their first-order neighbors. However, the local relative entropy~\cite{zhang2018measure} is directly measured by KL divergence whose value ranges in $[0, +\infty]$, and thus the entropy has no practical meaning when the value is too large. Therefore, we optimize it by introducing Jensen–Shannon (JS) divergence~\cite{lin1991divergence} with values in the range $[0,1]$ to measure semantically related remote nodes more accurately and more reasonable.

According to the definition of JS divergence, we first compute the KL divergence $D_{KL}(v\Vert \frac{v+u}{2})$ from a node $u$ to the node $v$ in a graph $G$, which is calculated as follows: 

\begin{flalign}
\label{equation6} d(v) &= [d_{v1}, d_{v2}, ..., d_{vM}] \in \mathbb{R}^{M} \\
p(v) &= [p_{v1}, p_{v2}, ..., p_{vM}] \in \mathbb{R}^{M}\nonumber\\
&= [\frac{d_{v1}}{\sum_{i=1}^M d_{vi}}, \frac{d_{v2}}{\sum_{i=1}^M d_{vi}}, ..., \frac{d_{vm}}{\sum_{i=1}^M d_{vi}}] 
\end{flalign}
\begin{equation} 
\label{equation8} D_{KL}(v\Vert \frac{v+u}{2}) = \sum_{i=1}^{M} p_{vi} \log{\frac{p_{vi}}{\frac{p_{vi}+p_{ui}}{2}}}
\end{equation} 
where $d(v)$ is a descending-order sequence containing degree values $d_{vi}$ of node $v$ and its one-hop neighbors; $M\text{=}max\left \{ degree(v)|v\in G\right \}$ is the max degree of a node in a graph $G$; $d_{vi}\text{=}0$ if $i > (degree(v)+1)$; $\sum_{i\text{=}1}^M d_{vi}$ is the sum of the sorted sequences $d(v)$; $p(v)$ is the normalized result of $d(v)$. 

Then, the structural entropy between nodes $v$ and $u$ can be calculated based on $D_{KL}(v\Vert \frac{v+u}{2})$ and $D_{KL}(u\Vert \frac{v+u}{2})$ as follows:

\begin{flalign} 
\label{equation9} H_s(v,u) &= 1- \frac{1}{2}\left(D_{KL}(v\Vert \frac{v+u}{2}) + D_{KL}(u\Vert \frac{v+u}{2})\right)
\end{flalign}
where $D_{KL}(v\Vert \frac{v+u}{2})$ represents the relative entropy from a node $u$ to the node $v$, and $D_{KL}(u\Vert \frac{v+u}{2})$ is the opposite; $H_s(v,u)$ is symmetric and denotes the structural entropy. As defined in Eq.~(\ref{equation9}), the larger value of the node structural entropy suggests the higher similarity between the nodes' structures.

\subsubsection{Node Relative Entropy.}
Node relative entropy consists of feature entropy and structural entropy. Formally, 

\begin{flalign} 
\label{equation1} H(v,u) &= H_f(v, u) + \lambda H_s(v,u)
\end{flalign}
where $H(v,u)$ represents node relative entropy between nodes $v$ and $u$; $H_f(v, u)$ is symmetric and denotes the feature entropy; $H_s(v,u)$ is symmetric and denotes the structural entropy; $\lambda$ is a hyper-parameter that controls the ratio. 

\subsubsection{Node Entropy Sequence Construction}
The original graph topology might not be appropriate for heterophilic GNNs with a message passing framework, where nodes of the same class have a high degree of structural similarity but may be far away from each other~\cite{zheng2022graph}. Considering the node relative entropy, which reflects the relevant relationship between nodes, a sorted sequence is constructed for each node according to the entropy value. Namely, $top\text{-}k$ new neighbors are selected for each node, and the original graph topology is optimized by connecting the selected new neighbors with the ego node. In this way, potential neighbors that are most similar to the ego node can be discovered and selected. Theoretically, since the node entropy sequence can be constructed flexibly to cover the whole graph, this strategy should be able to capture long-range dependencies even if the two nodes are distant. In this study, we select the $top\text{-}k$ nodes from the ego node's remote neighbors and delete the $top\text{-}d$ nodes from the one-hop neighbors based on the relative entropy. The values of $k$ and $d$ can be different on each individual node and they are updated by a deep reinforcement learning-based mechanism during model training.

\noindent \textbf{Time Complexity Analysis.} In the worst case, the time complexity of node relative entropy calculation is $O(N^2)$ for any size of the graph due to the matrix multiplication. By using PyTorch~\cite{paszke2017automatic} functions and matrix operations, we can significantly reduce the complexity of the calculation. For instance, the entropy is symmetric, and calculating the entropy between nodes $v$ and $u$ only needs to be performed once. Additionally, we only need to calculate the entropy once before training the GNN and DRL instead of in each epoch. In practice, graphs are usually sparse, and hence the empirical complexity is much smaller than $O(N^2)$ by using sparse matrix computation techniques.

\subsection{Deep Reinforcement Learning Module~\label{sec:drl}}

\begin{figure}[!t]
	\centering 
	\includegraphics[width=\linewidth]{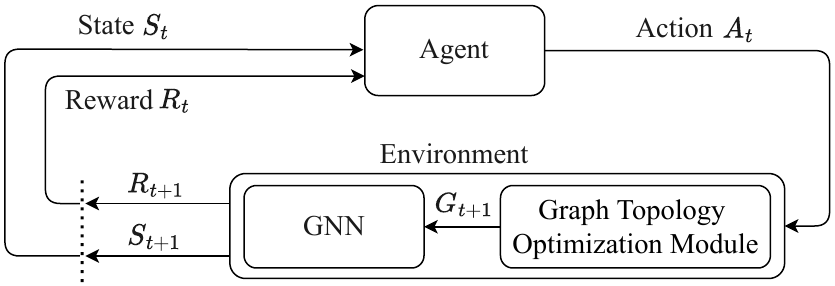} 
	\caption{The agent environment interaction in RL.} 
	\label{fig:DRL}
\end{figure}

\begin{figure}[!t]
	\centering 
	\includegraphics[width=\linewidth]{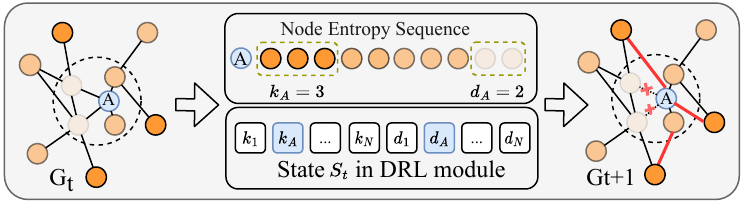} 
	\caption{An illustration of the graph topology optimization module. For any node $v$ in the graph $G_t$ at step $t$, the graph optimization module generates graph $G_{t+1}$ via adding $k_v$ edges and removing $d_v$ edges for node $A$ based on the state $S_t$. For node $A$ in this example, $k_A=3$ edges are added and $d_A=2$ edges are removed. } 
	\label{fig:GraphOptimization}
\end{figure}

To mitigate the ``personality of nodes" challenge in the $top\text{-}k$ and $top\text{-}d$ selection, this study proposes a deep reinforcement learning-based mechanism to update the $k$ and $d$ value of a node $v$ in a graph $G$ regardless of the size and topology structure of a graph. If $k$ (or $d$) is set as a hyper-parameter, meaning that all nodes will select the same number of $top\text{-}k$ new neighbors (or $top\text{-}d$  one-hop neighbors deleted), this can either introduce noisy nodes or discard informative nodes. Thus, instead of defining it as a single hyper-parameter, a DRL-based algorithm is used to find the optimal $top\text{-}k$ and $top\text{-}d$ values for every graph node. The updating process of $top\text{-}k$ and $top\text{-}d$ is modeled as a finite horizon Markov Decision Process (MDP), which is characterized by a tuple $(\mathcal{S}, \mathcal{A}, P, R, \gamma)$, where $S_t\in\mathcal{S}$, $A_t\in\mathcal{A}$ are state and action
observed at step $t$, $P$ is a state transition function, $R$ is a reward function, and $\gamma$ is a discount factor. The MDP tuple is defined as follows:
\begin{itemize}
\item {\textbf{State:} The state $S$ is set as $S\text{=}[k_1, k_2, ..., k_N, d_1, d_2, ..., d_N]$, where $k_i$ represents the number of newly connected neighbors of a node $v_i$, and $d_i$ is the number of deleted neighbors. The state at step $t$ is $S_t$; when $t\text{=}0$, $S_0\text{=}[0,0,...,0]$.}
\item {\textbf{Action:} The DRL agent updates $k_t$ by taking an action $A_t$ based on $S_{t}$. Since state $S_t$ is a multi-discrete state, action $A\text{=}[a^k_1,a^k_2,...,a^k_N,a^d_1,a^d_2,...,a^d_N]$ is defined as a multi-discrete action, and $a^k_i$ ($a^d_i$) is defined as the number of newly connected neighbors $k_i$ (deleted neighbors $d_i$) added or subtracted by a fixed value $\Delta k\text{=}1$ or kept unchanged.}
\item {\textbf{State Transition:} The state transition function is defined by:

\begin{flalign} 
\label{equation:state} S_{t+1}&= P(A_t | S_t) = S_t+A_t
\end{flalign}

After updating the state $S$, the original graph topology is reconstructed by adding edges between node $v$ and its entropy sequence's $top\text{-}k_v$ nodes. 

}
\item {\textbf{Reward:} The goal of DRL is to find an optimal graph topology for the original graph and GNN model. A discrete reward function is defined based on the GNN performance as follows:

\begin{flalign} 
\label{equation10} R(S_t)&= (acc_{t}-acc_{t-1}) + \lambda_{r} (loss_{t-1}-loss_{t})
\end{flalign}
where $acc_{t}$ and $loss_{t}$ are node classification accuracy and loss of the GNN on the training set at step $t$, respectively; $\lambda_{r}$ is a hyper-parameter that controls the ratio. 
The reward function indicates that if the performance of GNN on graph $G_t$ is better than that on graph $G_{t-1}$, then the topology of $G_t$ is better than $G_{t-1}$. It is worth noting that the current reward function can be replaced with alternative ones as long as they are able to guide the DRL towards optimizing the original topology.
}

\item { \textbf{Graph Topology Optimization Module:} 
At every step $t$, the proposed method optimizes the graph topology from $G_t$ to $G_{t+1}$. The DRL policy network first generates an action $A_t$ based on the state $S_t$ at step $t$, and the environment updates the state to $S_{t+1}$ according to Eq.~(\ref{equation:state}). Then, the topology optimization module connects node $v$ and the $top\text{-}k_v$ nodes in the node entropy sequence $seq_v$ for each node $v \in V$ to construct graph $G_{t+1}$ (see Figure~\ref{fig:GraphOptimization}). Finally, the GNN model is trained on the new graph $G_{t+1}$ and provides its accuracy and loss value on the training dataset to the DRL module as the reward according to Eq.~(\ref{equation10}).
}

\end{itemize}

\begin{algorithm}[!t]
\caption{The overall process of GraphRARE}
\label{alg:algorithm}
\begin{algorithmic}[1] 
\REQUIRE
$G=(V,E,X,A)$ 
\ENSURE
Node label $y_v$

\alglinenonumber{// Calculate node relative entropy} 
\FOR{$v, u \in \mathbb{V}$}
\STATE Calculate feature entropy $H_f(v,u)$ via Eq.~(\ref{equation5})
\STATE Calculate structural entropy $H_s(v,u)$ via Eq.~(\ref{equation9})
\STATE Calculate relative entropy $H(v,u)$ via Eq.~(\ref{equation1})
\ENDFOR
\STATE Construct node entropy sequence based on $H(v,u)$

\STATE Initialize $max\_acc = 0$
\alglinenonumber{// Train GNN and DRL jointly in an end-to-end manner}
\FOR{iteration=1,2,...}
\alglinenonumber{// Evaluate on training set, no backward}
\STATE $acc_t, loss_t = GNN(G_t, X)$
\IF{{$acc_t > max\_acc$}}
\alglinenonumber{// Additional training on the current graph $G_t$ as it leads to a higher accuracy}
\STATE {$max\_acc \gets acc_t$}
\STATE {Train $GNN(G_t, X)$ for a few more epochs}
\ENDIF

\alglinenonumber{// DRL process}
\STATE $reward(S_t) \gets$ Eq.~(\ref{equation10})
\STATE $A_t\gets$ DRL policy network
\STATE $S_{t+1}=P(A_t | S_t)$ via Eq.~(\ref{equation:state})
\STATE Reconstruct graph $G_{t+1}$ based on $G_t$, $S_{t+1}$, and node entropy sequence
\ENDFOR
\end{algorithmic}
\end{algorithm}

During the training process at step $t$, the DRL agent and GNN model are trained jointly in an end-to-end manner to optimize the original graph topology dynamically. Figure~\ref{fig:DRL} illustrates the agent-environment interaction in reinforcement learning. 
{Firstly, the GNN model is evaluated on the training set based on the graph topology $G_t$, and calculate the accuracy $acc_t$ and $loss_t$. Notably, during this evaluation phase, no updates are made to the parameters of GNN. If $acc_t$ surpasses the previous best, we train the GNN model for a few epochs on the graph topology $G_t$ to complete the node classification task. To prevent overfitting on $G_t$, an early stopping strategy is implemented.} Secondly, the DRL agent calculates the reward according to Eq.~\ref{equation10} and updates the policy network. Thirdly, the DRL agent takes action $A_t$ and updates the state to $S_{t+1}$ according to Eq.~\ref{equation:state}. Lastly, the graph topology optimization module reconstructs the graph topology by connecting the selected remote nodes based on the state $S_{t+1}$.
{The complete process is shown in Lines 7$-$18 in Algorithm~\ref{alg:algorithm}.} 

Lastly, the graph topology optimization module reconstructs the graph topology by connecting the selected remote nodes based on the state $S_{t+1}$. 

As the DRL algorithm, we use the Proximal Policy Optimization algorithm (PPO)~\cite{schulman2017proximal}, which is a classical policy gradient optimization algorithm that can deal with large-scale multi-discrete states and action spaces. It should be noted that in addition to the PPO algorithm, other reinforcement learning algorithms can also be conveniently applied to the proposed framework.

\subsection{Node Feature Learning\label{sec:nfl}}
The GraphRARE framework can be easily adapted to any existing GNN model to improve the model's performance. In the proposed framework, the original graph topology is reconstructed by training the DRL and GNN jointly based on the node relative entropy. In order to retain the GNN advantages, we adopt the same mechanism of node feature learning in the existing GNNs for multi-layer local aggregation. A GNN model consists of $L$ local aggregation layers, allowing each node to access the $L$-hop neighbors' information. For a node $v\in V$, the calculation formula of the $l$th layer in a $L$-layer GNN ($l\text{=}1,2,...,L$) is given by:

\begin{flalign} 
\label{equation11} m_v^{(l)} &= \text{AGGREGATE}^{(l)}\left (\left\{ h_u^{(l-1)}:u\in \mathcal N_1(v) \right\} \right)\\
h_v^{(l)} &=\text{UPDATE}^{(l)} \left (h_v^{(l-1)}, m_v^{(l)} \right)
\end{flalign}where $h_v^{(l)}$ is the feature of a node $v$ in the $l$th layer, and $h_v^{(0)}\text{=}x_v$; $\text{AGGREGATE}^{(l)}(\cdot)$ and $\text{UPDATE}^{(l)}(\cdot)$ represent the feature aggregation function (e.g., mean, LSTM, and max pooling) and feature update function (e.g., linear-layer combination and MLP)~\cite{zheng2022graph}, respectively; $\mathcal N_1(v)$ represents the one-hop neighboring node set of a node $v$.

\subsection{Model Training}
In Algorithm~\ref{alg:algorithm}, we present a step-by-step depiction of the training process. Beginning with a heterophilic graph $G=(V,E,X,A)$, we compute the relative entropy between pairs of nodes, considering both their node features and local structures (lines 1-5). Subsequently, we construct entropy sequences for each node in descending order, based on their relative entropy values (line 6). Lastly, we perform joint training of GNN and DRL module in an end-to-end manner. During iteration $t$, the GNN evaluates the graph $G_t$ and produces accuracy and loss metrics, which are then utilized to calculate the reward. The DRL policy network is updated based on the reward, and we  optimize the graph to $G_{t+1}$ for the subsequent iteration.

\section{Experiments\label{sec:exp}}
In this section, we evaluate the performance of the GraphRARE framework on the node classification task on seven real-world datasets. We first introduce the datasets, the baseline models, and hyper-parameter settings. Then, we compare GraphRARE with baseline models and analyze the results. Finally, the contribution of different components to the performance of GraphRARE is evaluated via hyper-parameter analysis, ablation study on the DRL module, convergence analysis for model training, and graph topology optimization analysis.  

\subsection{Datasets}

The proposed GraphRARE is evaluated on seven widely used real-world graph datasets, including five heterophilic graph datasets and two homophilic graph datasets. The heterophilic graph datasets are Chameleon, Squirrel~\cite{DBLP:journals/compnet/RozemberczkiAS21}, Cornell, Texas, and Wisconsin~\cite{DBLP:journals/tfs/Garcia-PlazaFMZ17}. The homophilic graph datasets are Cora~\cite{sen2008collective} and Pubmed~\cite{namata2012query}. The statistics for all the datasets are summarized in Table~\ref{table:datasets}. 

\begin{table}[!t]
    \caption{Statistics and properties of the seven datasets.}
    \renewcommand\arraystretch{1.01}
    \centering
    \scalebox{1}{
    \begin{tabular}{lccccc}
    \toprule
    \textbf{Datasets} & \#Nodes & \#Edges & \#Features & \#Classes & $\mathcal{H}$\\
    \midrule
    \textbf{Chameleon} & 2,277 & 36,101 & 2,325 & 5 & 0.23 \\
    \textbf{Squirrel}  & 5,201 &217,073 & 2,089 & 5 & 0.22 \\
    \textbf{Cornell}   & 183   & 295    & 1,703 & 5 & 0.30 \\
    \textbf{Texas}     & 183   & 309    & 1,703 & 5 & 0.11 \\
    \textbf{Wisconsin} & 251   & 499    & 1,703 & 5 & 0.21 \\
    \textbf{Cora}      & 2,708 & 5,429  & 1,433 & 7 & 0.81 \\
    \textbf{Pubmed}    & 19,717& 44,338 &  500  & 3 & 0.80 \\
    
    \bottomrule
    \end{tabular}
    }
    \label{table:datasets}
\end{table}

\subsection{Compared Methods and Setup}
In the experiments, four enhanced GNNs based on the proposed framework are constructed: GCN-RARE, GraphSAGE-RARE, GAT-RARE, and H2GCN-RARE. To be specific, GCN-RARE denotes reinforcement learning enhanced GCN model with relative entropy, and so do the other three models. The models are end-to-end trained using the node classification loss. 
{The proposed four enhanced GNNs are compared with thirteen baseline models, including an attributed-only based MLP, three traditional GNN models (GCN, GraphSAGE, and GAT), and nine GNN-based state-of-the-art (SOTA) methods for heterophilic graphs (MixHop, H2GCN, Geom-GCN, UGCN, SimP-GCN, OTGNet, GBK-GNN, Polar-GNN, and HOG-GCN) for node classification tasks. Specifically, the OTGNet model originally handles the heterophily for temporal graph. Since the underlying mechanisms of OTGNet for managing diverse classes and propagating information are applicable across temporal and static graph structures, we adopt the static graphs as input for a fair comparison.} 

\begin{table*}[!t]
    \caption{{Mean accuracy and standard deviation over ten different data splits on the seven real-world graph datasets. The best results are highlighted in \textbf{bold}. The improvements of the GraphRARE models compared to their counterparts are calculated, with an improvement denoted by an up arrow $\uparrow$, and a negative improvement by a down arrow $\downarrow$.}}
    \renewcommand\arraystretch{1.0} 
    \centering
    \scalebox{1}{%
    \begin{tabular}{lcccccccccc}
    \toprule
    \textbf{Method} & \textbf{Chameleon} & \textbf{Squirrel} & \textbf{Cornell} & \textbf{Texas} & \textbf{Wisconsin} & \textbf{Cora} &  \textbf{Pubmed} & \textit{Average}\\
    \midrule
    MLP & 46.51{$\pm$2.53} & 29.29{$\pm$1.40} &  80.81{$\pm$6.91} & 81.08{$\pm$5.41} & 84.12{$\pm$2.69} & 74.61{$\pm$1.97} & 86.63{$\pm$0.38} & 69.01  \\
    \midrule
    GCN~\cite{DBLP:conf/iclr/KipfW17} & 59.08{$\pm$2.47} & 46.64{$\pm$1.42} & 55.73{$\pm$6.33} & 52.84{$\pm$5.43} & 56.04{$\pm$5.76} & 85.16{$\pm$1.01} & 87.18{$\pm$0.42} & 63.24  \\
    GraphSAGE~\cite{DBLP:conf/nips/HamiltonYL17} & 58.83{$\pm$2.15} & 41.44{$\pm$1.85} & 72.70{$\pm$9.16} & 75.68{$\pm$6.16} & 76.08{$\pm$4.28} & 84.53{$\pm$1.38} & 85.09{$\pm$0.52} &  70.62\\
    GAT~\cite{DBLP:conf/iclr/VelickovicCCRLB18} & 54.34{$\pm$2.46} & 40.79{$\pm$3.78} &  54.22{$\pm$5.38} & 56.49{$\pm$6.89} & 54.45{$\pm$5.62} & 86.02{$\pm$1.37} & 86.55{$\pm$0.47}&  61.84 \\
    \midrule
    MixHop~\cite{abu2019mixhop} & 60.50{$\pm$2.53} & 43.80{$\pm$1.48} &  73.51{$\pm$6.34} & 77.84{$\pm$7.73} & 75.88{$\pm$4.90} & 83.10{$\pm$2.03} & 80.75{$\pm$2.29} &  70.77\\
    H2GCN~\cite{DBLP:conf/nips/ZhuYZHAK20} & 56.85{$\pm$1.68} & 32.20{$\pm$2.19} &  78.16{$\pm$4.05} & 79.70{$\pm$5.16} & 82.08{$\pm$3.22} & 86.26{$\pm$1.08} & 88.76{$\pm$0.41} & 72.00\\
    Geom-GCN~\cite{DBLP:conf/iclr/PeiWCLY20} & 60.90 & 38.14 & 60.81 & 67.57 & 64.12 & 85.27 & 90.05 &  66.69\\
    UGCN~\cite{DBLP:conf/nips/JinYHWWHH21} & 54.07 & 34.39 &  69.77 & 71.72 & 69.89 & 84.00 & 85.22 &  67.01\\
    SimP-GCN~\cite{DBLP:conf/wsdm/JinDW0LT21} & 62.61 & 42.57 &  84.05 & 81.62 & 85.49 & 82.80 & 81.10 &  74.33\\
    {OTGNet~\cite{feng2023towards}} & {46.34$\pm$5.77} & {35.39$\pm$2.96} & {58.19$\pm$9.61} & {65.81$\pm$10.38} & {61.23$\pm$8.69} & {73.31$\pm$4.48} & {76.64$\pm$2.18} & {59.56}\\
    
    {GBK-GNN~\cite{DBLP:conf/www/DuSFMLHZ22}} & {48.46{$\pm$0.81}} & {36.69{$\pm$0.33}} & {69.59{$\pm$2.24}} & {75.59{$\pm$2.94}} & {78.98{$\pm$4.19}} & {82.65{$\pm$0.64}} & {83.48{$\pm$0.19}} & {67.92}\\
    {Polar-GNN~\cite{DBLP:conf/www/FangXSLZ22}} & {64.0{$\pm$0.6}} & {49.3{$\pm$0.8}} & {-} & {-} & {-} & {83.1{$\pm$0.9}} & {80.2{$\pm$0.4}} & {-} \\
    HOG-GCN~\cite{DBLP:conf/aaai/WangJWHH22} & 54.01{$\pm$1.28} & 35.46{$\pm$1.96} &  84.32{$\pm$4.32} & 85.17{$\pm$4.40} & 86.67{$\pm$3.36} & 87.04{$\pm$1.10} & 88.79{$\pm$0.40} &  74.49 \\
    \midrule 
    
    \multirow{2}{*}{\textbf{GCN-RARE}(ours)} & 68.05{$\pm$1.87} & \textbf{55.90}{$\pm$1.39} &  64.59{$\pm$4.95} & 58.38{$\pm$6.64} & 61.76{$\pm$5.49} & \textbf{87.24}{$\pm$1.26} & 88.41{$\pm$0.50} &  69.19 \\
      & $\uparrow$8.97 & $\uparrow$9.26 & $\uparrow$8.86 & $\uparrow$5.54 & $\uparrow$5.72 & $\uparrow$2.08 & $\uparrow$1.23 & $\uparrow$5.95\\
    \midrule 
    \multirow{2}{*}{\textbf{GraphSAGE-RARE}(ours)} & \textbf{69.28}{$\pm$1.90} & 52.84$\pm$1.33 &  82.97{$\pm$5.10} & 82.16{$\pm$6.07} & 85.69{$\pm$5.29} & {87.08}{$\pm$1.17} & 89.03{$\pm$0.56} &  \textbf{78.43} \\
      & $\uparrow$10.45 & $\uparrow$11.40 & $\uparrow$10.27 & $\uparrow$6.48 & $\uparrow$9.61 & $\uparrow$2.55 & $\uparrow$3.94 & $\uparrow$7.81\\
    \midrule 
    \multirow{2}{*}{\textbf{GAT-RARE}(ours)} & 64.56{$\pm$2.48} & 49.99{$\pm$2.79}  & {61.60{$\pm$3.01}} & 58.11{$\pm$5.44} & 61.08{$\pm$3.80} & 86.60{$\pm$1.19} & 87.41{$\pm$0.65} & {67.05} \\
      & $\uparrow$10.22 & $\uparrow$9.20 & {$\uparrow$7.38} & $\uparrow$1.62 & $\uparrow$6.63 & $\uparrow$0.58 & $\uparrow$0.86 & {$\uparrow$5.21}\\
    \midrule 
    \multirow{2}{*}{\textbf{H2GCN-RARE}(ours)} & 58.09{$\pm$1.91} & {34.93{$\pm$1.49}} & \textbf{87.84}{$\pm$4.05} & \textbf{86.76}{$\pm$5.80} & \textbf{90.00}{$\pm$2.97} & {86.82}{$\pm$1.51} & \textbf{90.07}{$\pm$0.26} &  {76.36} \\
      & $\uparrow$1.24 & {$\uparrow$2.73} & $\uparrow$9.68 & $\uparrow$7.06 & $\uparrow$7.92 & $\uparrow$0.56 & $\uparrow$1.31 & {$\uparrow$4.36}\\
    \bottomrule
    \end{tabular}
    }
    \label{table:experiment}
\end{table*}

\subsection{Hyper-parameter setting}
Based on the hyper-parameter settings of the original models~\cite{DBLP:conf/iclr/KipfW17, DBLP:conf/nips/HamiltonYL17, DBLP:conf/iclr/VelickovicCCRLB18, DBLP:conf/nips/ZhuYZHAK20}, the final hyper-parameter setting is configured as follows: the dropout rate is set to $p\text{=}0.5$, the initial learning rate is 0.05, and the weight decay is set to $\{5E-5, 5E-6\}$. For GCN, GAT, GraphSAGE, and H2GCN, the number of hidden units is selected from the set \{48, 64, 128\}, the number of GNN layers is set to two, and the Adam optimizer is adopted. For the sake of fair comparison, in GCN-RARE, GraphSAGE-RARE, GAT-RARE, and H2GCN-RARE, the hyper-parameters are configured to be the same as those in their counterparts (GCN, GraphSAGE, GAT, and H2GCN). We set the $\lambda$ in Eq.(\ref{equation1}) to $\lambda\text{=}1.0$ and discuss it in Section~\ref{section:hyperparameter}. For the PPO algorithm, we choose a MLP network for the policy model.

For all the benchmarks, we adopt the same class labels, feature vectors, and perform 10 random splits on the datasets (60\%/20\%/20\% of nodes per class for training/validation/testing) provided by Pei et al.~\cite{DBLP:conf/iclr/PeiWCLY20}, which is available on their GitHub~\footnote{https://github.com/graphdml-uiuc-jlu/geom-gcn/tree/master/splits}. We launch the testing procedure when the validation accuracy of the trained model achieves a maximum value for each run, and calculate the average test accuracy over these ten runs.

{
The GCN, GraphSAGE, GAT, H2GCN and our methods are constructed using Pytorch~\cite{paszke2017automatic}, Deep Graph Library~\cite{wang2019deep}, Pytorch Geometric~\cite{Fey/Lenssen/2019}, OpenAI Gym~\cite{gym}, and Stable-Baselines3~\cite{stable-baselines3} with one NVIDIA A100-40GB GPU.}

\begin{table*}[!t]
    \caption{Hyper-parameter analysis results. The hyper-parameter $\lambda$ controls the weight of structural entropy in the node relative entropy as defined in~Eq.(\ref{equation1}). The best results are highlighted in \textbf{bold}.}
    \renewcommand\arraystretch{1.0}
    \centering
    \scalebox{1}{
    \begin{tabular}{lccccccccccc}
    \toprule
    \textbf{Method} & \textbf{$\lambda$} & \textbf{Chameleon} & \textbf{Squirrel} & \textbf{Cornell} & \textbf{Texas} & \textbf{Wisconsin} & \textbf{Cora} & \textbf{Pubmed} & \textit{Average}\\
    \midrule
    
    \multirow{4}{*}{\textbf{GCN-RARE}} 
    & 0.1 & 67.36{$\pm$2.25} & 54.89{$\pm$1.54} & 63.92{$\pm$7.24} & 57.83{$\pm$7.81} & 59.31{$\pm$5.07} & \textbf{87.34}{$\pm$1.07} & 87.49{$\pm$0.72} & 68.31 \\
    & 0.5 & 67.56{$\pm$2.01} & 54.77{$\pm$1.46} & 63.77{$\pm$5.69} & 57.78{$\pm$6.36} & 58.93{$\pm$4.81} & 86.21{$\pm$1.33} & 87.62{$\pm$0.74} & 68.10 \\
    & 1.0 & \textbf{68.05}{$\pm$1.87} & \textbf{55.90}{$\pm$1.39} &  \textbf{64.59}{$\pm$4.95} & \textbf{58.38}{$\pm$6.64} & \textbf{61.76}{$\pm$5.49} & 87.24{$\pm$1.26} & \textbf{88.41}{$\pm$0.50} &  \textbf{69.19} \\
    & 10.0 & 67.73{$\pm$1.98} & 55.45{$\pm$1.25} & 63.54{$\pm$6.23} & 57.79{$\pm$7.46} & 58.82{$\pm$4.60} & 86.27{$\pm$1.48} & 87.77{$\pm$0.97} & 68.20 \\
    \midrule
    
    \multirow{4}{*}{\textbf{GraphSAGE-RARE}} 
    & 0.1 & 68.70{$\pm$2.66} & \textbf{52.93}{$\pm$1.74} & 80.03{$\pm$5.70} & 80.54{$\pm$5.90} & 84.16{$\pm$4.68} & 86.74{$\pm$1.37} & 88.40{$\pm$0.74} & 77.36 \\
    & 0.5 & 68.95{$\pm$2.37} & 52.58{$\pm$1.61} & 81.00{$\pm$8.48} & 81.46{$\pm$5.57} & 84.94{$\pm$3.73} & 87.06{$\pm$0.82} & \textbf{89.12}{$\pm$0.38} & 77.87 \\
    & 1.0 & \textbf{69.28}{$\pm$1.90} & 52.84{$\pm$1.33} & \textbf{82.97}{$\pm$5.10} & \textbf{82.16}{$\pm$6.07} & \textbf{85.69}{$\pm$5.29} & \textbf{87.08}{$\pm$1.17} & 89.03{$\pm$0.56} &  \textbf{78.43} \\
    & 10.0 & 68.78{$\pm$2.01} & 52.67{$\pm$1.41} & 81.54{$\pm$7.23} & 81.49{$\pm$5.14} & 84.96{$\pm$4.62} & 86.76{$\pm$1.33} & 88.77{$\pm$0.93} & 77.85 \\
    \midrule
    
    \multirow{4}{*}{\textbf{GAT-RARE}} 
    & 0.1 & 63.77{$\pm$2.32} & 49.54{$\pm$2.55} & 60.00{$\pm$7.03} & 56.89{$\pm$6.14} & 60.12{$\pm$6.15} & 86.22{$\pm$1.26} & 86.99{$\pm$0.83} & 66.22 \\
    & 0.5 & 63.18{$\pm$2.39} & 48.88{$\pm$2.30} & 60.84{$\pm$4.86} & 57.08{$\pm$5.47} & 60.71{$\pm$4.59} & 86.54{$\pm$1.29} & 87.38{$\pm$0.43} & 66.37 \\
    & 1.0 & \textbf{64.56}{$\pm$2.48} & \textbf{49.99}{$\pm$2.79}  & \textbf{61.60}{$\pm$3.01} & \textbf{58.11}{$\pm$5.44} & \textbf{61.08}{$\pm$3.80} & \textbf{86.60}{$\pm$1.19} & \textbf{87.41}{$\pm$0.65} & \textbf{67.05} \\
    & 10.0 & 64.30{$\pm$3.04} & 49.96{$\pm$2.25} & 59.73{$\pm$4.75} & 56.24{$\pm$6.40} & 60.90{$\pm$6.74} & 86.12{$\pm$1.43} & 87.16{$\pm$0.57} & 66.34 \\
    \midrule
    
    \multirow{4}{*}{\textbf{H2GCN-RARE}} 
    & 0.1 & 58.03{$\pm$1.64} & 33.77{$\pm$2.19} & 86.86{$\pm$5.37} & 86.31{$\pm$3.60} & 88.43{$\pm$4.15} & 86.40{$\pm$1.24} & 89.48{$\pm$1.24} & 75.61 \\
    & 0.5 & 57.70{$\pm$1.66} & 33.65{$\pm$2.03} & 87.42{$\pm$5.92} & 86.12{$\pm$3.51} & 87.65{$\pm$3.17} & 86.08{$\pm$1.76} & 89.67{$\pm$0.59} & 75.47 \\
    & 1.0 & \textbf{58.09}{$\pm$1.91} & \textbf{34.93}{$\pm$1.49} & \textbf{87.84}{$\pm$4.05} & \textbf{86.76}{$\pm$5.80} & \textbf{90.00}{$\pm$2.97} & \textbf{86.82}{$\pm$1.51} & \textbf{90.07}{$\pm$0.26} &  \textbf{76.36} \\
    & 10.0 & 57.35{$\pm$1.74} & 32.90{$\pm$1.83} & 86.24{$\pm$5.51} & 85.86{$\pm$3.67} & 87.65{$\pm$3.17} & \textbf{86.82}{$\pm$1.42} & 89.35{$\pm$0.92} & 75.17 \\
    \bottomrule
    \end{tabular}
    }

    \label{table:hyperparameter}
\end{table*}

\begin{table*}[!t]
    \caption{{Ablation study on relative entropy and DRL module. ``GCN-RE[$\cdot$]" represents GraphRARE without DRL, instead randomly assigning $k$ and $d$ to each node. ``GCN-RA" denotes GraphRARE without the relative entropy component. ``GCN-RARE-add" and ``GCN-RARE-remove" describe GraphRARE when it only adds edges and only removes edges, respectively.}}
    \renewcommand\arraystretch{1.0} 
    \centering
    \scalebox{1}{%
    \begin{tabular}{lcccccccccc}
    \toprule
    \textbf{Method} & \textbf{Chameleon} & \textbf{Squirrel} & \textbf{Cornell} & \textbf{Texas} & \textbf{Wisconsin} & \textbf{Cora} &  \textbf{Pubmed} & \textit{Average}\\
    \midrule
    GCN~\cite{DBLP:conf/iclr/KipfW17} & 59.08{$\pm$2.47} & 46.64{$\pm$1.42} & 55.73{$\pm$6.33} & 52.84{$\pm$5.43} & 56.04{$\pm$5.76} & 85.16{$\pm$1.01} & 87.18{$\pm$0.42} & 63.24  \\
    \midrule 
    \textbf{GCN-RE[0..5]}(ours) & {63.48$\pm$2.09} & {48.03$\pm$1.05} & {59.72$\pm$3.72} & {55.43$\pm$9.48} & {56.17$\pm$4.38} & {84.32$\pm$1.60} & {85.13$\pm$0.46} & 64.61 \\
    \textbf{GCN-RE[0..10]}(ours) & {60.89$\pm$2.31} & {46.04$\pm$0.73} & {61.35$\pm$3.64} & {56.21$\pm$5.75} & {59.49$\pm$5.33} & {83.44$\pm$1.59} & {84.52$\pm$0.39} & 64.56 \\
    \textbf{GCN-RE[0..15]}(ours) & {58.69$\pm$2.16} & {45.19$\pm$1.37} & {61.08$\pm$5.01} & {53.74$\pm$6.13} & {61.25$\pm$5.04} & {83.11$\pm$1.63} & {84.15$\pm$0.40} & 63.89 \\
    \textbf{GCN-RE[0..20]}(ours) & {58.55$\pm$1.40} & {45.09$\pm$1.42} & {61.05$\pm$7.57} & {58.61$\pm$7.92} & {59.62$\pm$7.37} & {83.27$\pm$1.36} & {83.97$\pm$0.49} & 64.31 \\
    \midrule
    \textbf{GCN-RA}(ours) & {61.48$\pm$1.39} & {47.50$\pm$1.71} & {59.57$\pm$6.56} & {54.57$\pm$7.94} & {59.65$\pm$7.83} & {84.98$\pm$1.37} & {87.42$\pm$0.59} & 63.78 \\
    \midrule
    \textbf{GCN-RARE-add}(ours) & {66.43$\pm$1.36} & {55.46$\pm$1.14} & {58.11$\pm$6.19} & {58.12$\pm$7.74} & {59.22$\pm$4.45} & {86.58$\pm$1.34} & {88.02$\pm$0.52} & 67.25 \\
    \textbf{GCN-RARE-remove}(ours) & {67.52$\pm$1.68} & {55.43$\pm$1.42} & {60.95$\pm$4.55} & {55.14$\pm$7.66} & {61.37$\pm$7.07} & {86.88$\pm$1.23} & {87.95$\pm$0.62} & 67.89 \\
    \midrule
    {\textbf{GCN-RARE-reward}(ours)} & {{66.54$\pm$2.12}} & {{53.05$\pm$1.33}} & {{60.64$\pm$4.99}} & {{54.02$\pm$7.26}} & {{58.74$\pm$6.25}} & {{86.72$\pm$1.37}} & {{87.74$\pm$0.57}} & {66.78} \\
    \midrule
    \textbf{GCN-RARE}(ours) & \textbf{68.05}{$\pm$1.87} & \textbf{55.90}{$\pm$1.39} &  \textbf{64.59}{$\pm$4.95} & \textbf{58.38}{$\pm$6.64} & \textbf{61.76}{$\pm$5.49} & \textbf{87.24}{$\pm$1.26} & \textbf{88.41}{$\pm$0.50} &  \textbf{69.19} \\
    \bottomrule
    \end{tabular}
    }
    \label{table:abaltion-exp}
\end{table*}

\subsection{Node Classification}
The experimental results for the node classification task on the seven datasets are illustrated in Table~\ref{table:experiment}. The mean accuracy and standard deviation for ten different data splits are presented, and the best results are highlighted in bold.

With the node relative entropy guided higher-order neighbor ranking and the deep reinforcement learning-based graph topology optimization, GraphRARE obtains a better graph topology under the heterophily setting. We apply different enhanced GraphRARE models and make comparisons with corresponding baselines and the GNN-based SOTA models. As displayed in Table~\ref{table:experiment}, three patterns can be observed: (1) The enhanced models, GCN-RARE, GraphSAGE-RARE, GAT-RARE, and H2GCN-RARE, demonstrate improvements over their counterparts (GCN, GraphSAGE, GAT, and H2GCN) on all the heterophilic datasets, which indicates that the advantages of the proposed GraphRARE framework are general for common GNNs. Specifically, the proposed framework improves the performance of GCN by 5.95\% on average, GraphSAGE by 7.81\% on average, GAT by 5.14\% on average, and H2GCN by 4.23\% on average. (2) On the dataset with strong homophily (Cora and Pubmed), GraphRARE performs better or is comparable to the baselines. To be specific, GraphRARE performs best on Cora, and achieves the second best on Pubmed. The experimental results show that the GraphRARE framework also performs well on homophilic graphs. (3) On the dataset with middle and low homophily, e.g., Chameleon, Squirrel, the four GraphRARE-based models achieve the best results on the five heterophilic datasets, and the overall performance is 6.69\% higher than the best SOTA method (SimP-GCN) on average. This is mainly attributed to the fact that GraphRARE adaptively optimizes the graph topology based on the DRL module and the node relative entropy, it effectively ignores the irrelevant nodes and selects the most important remote nodes by considering the node feature and structural information. 

\subsection{Hyper-parameter Analysis} 
\label{section:hyperparameter} 
With the calculation of the node relative entropy, the GraphRARE framework optimizes the graph topology adaptively. As defined in~Eq.(\ref{equation1}), the hyper-parameter $\lambda$ controls the weight of structural entropy in the node relative entropy. In order to determine the weight of the structure entropy, we conduct extensive experiments to analyze $\lambda$ on the benchmarks. The experimental results are depicted in Table~\ref{table:hyperparameter}. As shown in the table, it is not hard to see that the structural entropy is often equally important as the feature entropy for the GraphRARE framework. Therefore, given an arbitrary graph, we can simply set $\lambda$ to the default value of 1.0. As the relative entropy with $\lambda\text{=}0.1$ or $\lambda\text{=}10.0$ can be considered as feature entropy alone or structural entropy alone respectively, we also observe that the performance is better when considering both feature entropy and structure entropy than when considering either of them separately. The experimental results verify our assumption that both the node features and the local topological structure contribute to measuring the mutual information between node pairs.

\begin{figure*}[!t]
	\centering 
	\includegraphics[width=\linewidth]{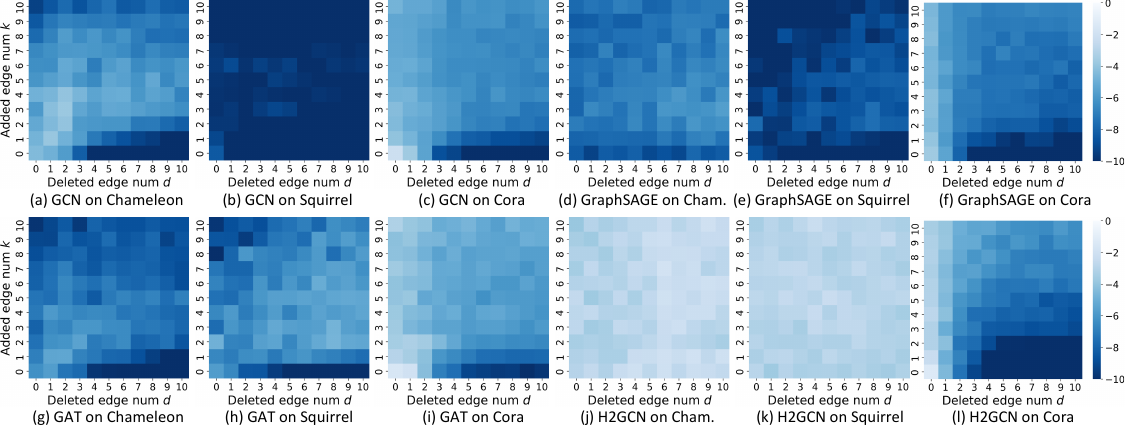} 
	\caption{Ablation study on the DRL module's contribution. Each heatmap compares GraphRARE's performance with and without the DRL module. The horizontal axis represents the number of deleted edges ($d$), while the vertical axis represents the number of added edges ($k$). Deeper colors indicate more significant performance degradation compared to GraphRARE.}
     \label{fig:ablationStudy}
\end{figure*}

\begin{figure*}[ht!]
     \centering
         \begin{subfigure}{0.32\textwidth}
             \centering
             \includegraphics[width=\textwidth]{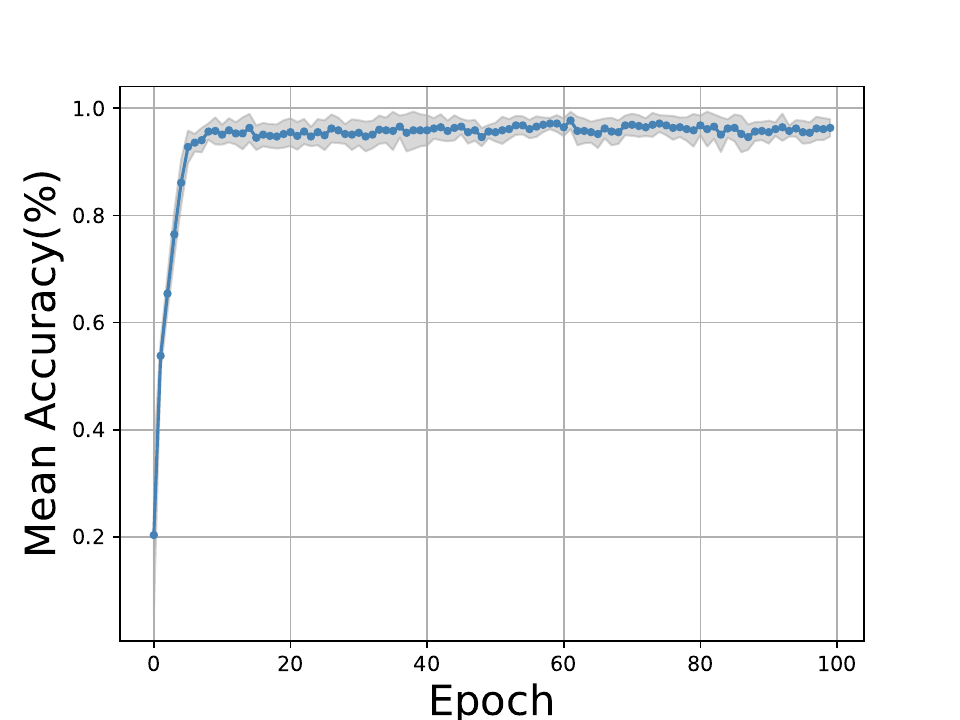}
             \caption{Training process of GraphRARE.}
             \label{fig:train_acc}
         \end{subfigure}
         \begin{subfigure}{0.32\textwidth}
             \centering
             \includegraphics[width=\textwidth]{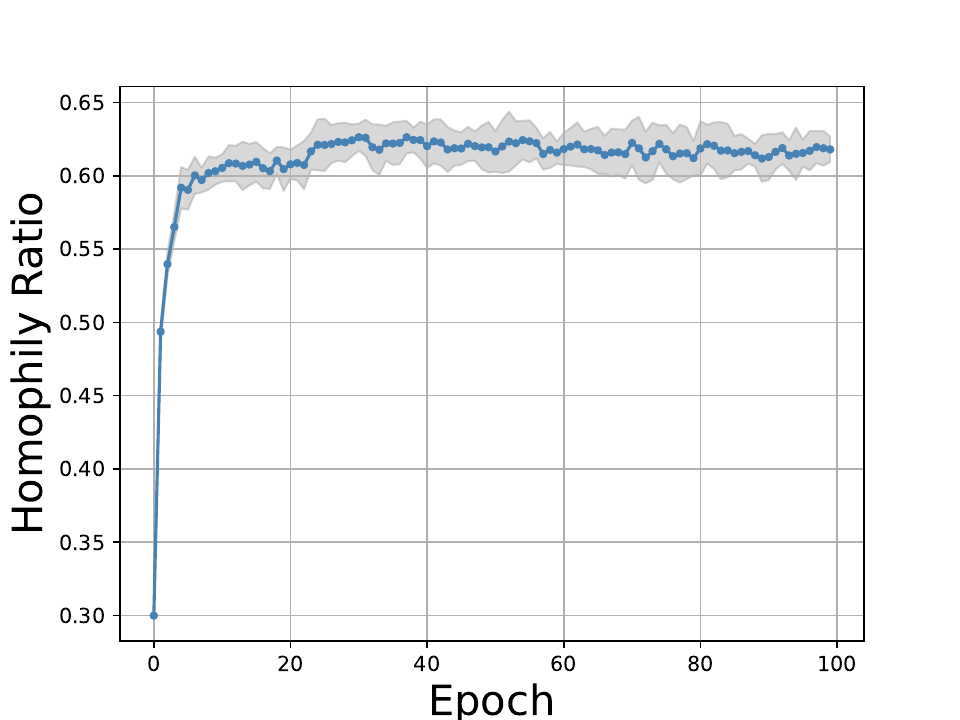}
             \caption{Updating process of homophily ratio.}
             \label{fig:train_homo}
        \end{subfigure}
         \begin{subfigure}{0.32\textwidth}
             \centering
             \includegraphics[width=\textwidth]{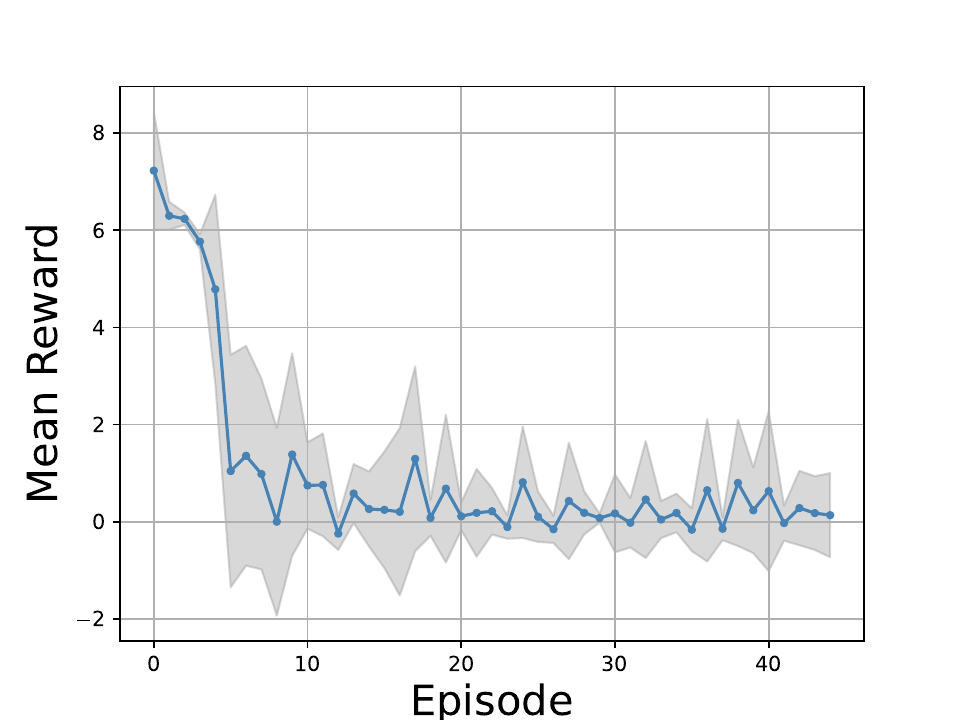}
             \caption{Learning curve of DRL.}
             \label{fig:train_reward}
         \end{subfigure}
        \caption{The training process of GraphRARE. The subfigures illustrate the training performance of GCN-RARE on the Cornell dataset. Subfigure~\ref{fig:train_acc} shows node classification accuracy, Subfigure~\ref{fig:train_homo} shows graph homophily ratio, and Subfigure~\ref{fig:train_reward} shows the mean reward of the DRL module in each episode.}
        \label{fig:train_process}
\end{figure*}

\subsection{Ablation Study}\label{sec:ablation_study}
The proposed GraphRARE framework jointly trains the DRL-based module and GNN to optimize the graph topology. An ablation study on the GraphRARE framework is conducted to evaluate the DRL module and the relative entropy's contribution by (1) setting the same number of new neighbors $k$ and deleted neighbors $d$ for each node to a fixed value. (2) randomly allocating different $k$ and $d$ value for ego nodes. (3) randomly shuffling the node entropy sequence for each node, which is GraphRARE without relative entropy. (4) only adding edges or removing edges in the original graphs. {(5) replace the reward function in DRL module.}

\subsubsection{Under the same $k$ and $d$ value setting}
To fully investigate the performance under different settings, we set the fixed new neighbor number $k$ from one to ten for the four enhanced GNNs on three datasets (Chameleon, Squirrel, and Cora). The experimental results are depicted in Figure~\ref{fig:ablationStudy}, and two patterns can be observed: (1) For the four backbone models on the three datasets, the proposed framework with the DRL module outperforms all of the models with a fixed new/deleted neighbor number. As discussed above, this could be because the model with a fixed $k$ and $d$ falls into a local sub-optimal solution due to the mixture of useful information and noises of remote nodes. (2) Removing edges has a greater impact on performance than adding edges in most experiments. When edges are removed, it can lead to the disconnection of nodes or subgraphs, reducing the information flow and the ability of GNNs to propagate messages effectively across the graph. 

\subsubsection{Under the random $k$ and $d$ value setting}
We conduct experiments for GCN on seven datasets that randomly allocate different k-value for ego nodes. The experimental results are illustrated in Table~\ref{table:abaltion-exp} (see GCN-RE), which shows that randomly allocating different k-values for nodes can improve the performance of the backbone GNN but is inferior to that of using DRL module in some datasets. The results demonstrate the effectiveness of both relative entropy and DRL module.

\subsubsection{Without the relative entropy}
The node entropy sequence is built based on the value of node relative entropy. To evaluate the contribution of the relative entropy, we randomly shuffle the node entropy sequence for each node, which equals ``GraphRARE without relative entropy". The results in Table~\ref{table:abaltion-exp} (see GCN-RA) emphasize the criticality of defining an appropriate metric for measuring the node importance when optimizing the graph topology.

\subsubsection{Without adding or removing edge operation}
In our framework, we jointly train the GNN and DRL module to optimize the graph topology by simultaneously adding and removing edges. The results in Table~\ref{table:abaltion-exp} (see GCN-RARE-add and GCN-RARE-remove) highlight the significance of adding informative new neighbors and removing noisy neighbors are both important.

\subsubsection{{Alternative reward function}}
{The reward function is crucial as it guides the policy towards optimized decisions. In order to explore the impact of different reward design choices on the final performance of our model, we replace the reward function in Eq.~\ref{equation10} with AUC score. The results in Table~\ref{table:abaltion-exp} (see GCN-RARE-reward) show that an appropriate reward function is beneficial for guiding the optimization process.}

\subsection{Efficiency Study}

To further show the efficiency of our GraphRARE, we compare our approach with baselines on the same device as shown in Table~\ref{table:efficiency}. Each model is trained for 500 epochs on each dataset and the average training time per epoch is reported. 
{Additionally, we report the computation time required for the node relative entropy calculation, which is computed only once before the model training. } The computational cost is comparable to or more efficient than SOTA models (Simp-GCN and HOG-GCN).

\begin{table}[!t]
    \caption{{Comparisons in terms of real running time. Those appended by an asterisk [$\ast$] denotes SOTA models. The relative entropy is only computed once before model training.}}
    \resizebox{\linewidth}{!}{
    \renewcommand\arraystretch{1.0} 
    \centering
    \begin{tabular}{lccccccc}
    \toprule
    \textbf{Method} & \textbf{Chameleon} & \textbf{Squirrel} & \textbf{Cornell} & \textbf{Texas} & \textbf{Wisconsin}\\
    \midrule
    
    GCN & 11.36 & 13.3 & 9.00 & 9.32 & 9.32 \\ 
    GAT & 34.10 & 57.16 & 21.52 & 20.68 & 21.90 \\
    GraphSAGE & 12.68 & 13 & 11.04 & 11.16 & 12.70 \\ 
    H2GCN & 25.52 & 57.46 & 13.58 & 16.18 & 15.62 \\ 
    \midrule
    SimP-GCN$\ast$ & 35.70 & 44.86 & 19.68 & 18.64 & 20.68 \\ 
    HOG-GCN$\ast$ & 77.28 & 246.60 & 56.46 & 55.05 & 53.34 \\ 
    \midrule
    GCN-RARE (ours) & 57.44 & 186.12 & 16.40 & 19.38 & 16.58 \\ 
    GAT-RARE (ours) & 66.34 & 209.88 & 33.70 & 26.98 & 25.77 \\ 
    GraphSAGE-RARE (ours) & 41.06 & 95.04 & 24.17 & 28.72 & 26.11 \\ 
    H2GCN-RARE (ours) & 70.61 & 229.07 & 22.04 & 25.09 & 31.29 \\
    \midrule

    {Entropy Computation} & {28.67} & {266.48} & {0.0596} & {0.0615} & {0.1974}  \\

    \bottomrule
    \end{tabular}
    \label{table:efficiency}
    }
\end{table}

\subsection{Convergence of GraphRARE}
Since the DRL algorithm and the GNN are trained jointly in the proposed GraphRARE framework, the updating and convergence process is indeed important. We visualize the node classification accuracy during model training, the average reward of reinforcement learning, and the homophily ratio of the graph topology as shown in Figure~\ref{fig:train_process}. In Figure~\ref{fig:train_acc}, the shadowed area is enclosed by the min and max value of ten cross-validation training runs, and the solid line in the middle is the mean value of each epoch. The result shows that the GNN model converges well when training with the DRL module under the GraphRARE framework. As shown in Figure~\ref{fig:train_reward}, the mean episode reward of the DRL module does not update very steadily at the beginning because the accuracy and loss of GNN change a lot. When the framework gradually converges, we can observe that the RL algorithm converges with a stable learning curve. Figure~\ref{fig:train_homo} shows the homophily ratio of the graph topology during the training process, the ratio convergences to the mean value of 0.63. The results show that the DRL module outputs a stable number of homogeneous adjacent nodes at the later stage of model training, and GNN also gives a stable classification result. In summary, in the early stage of model training, the accuracy of GNN rises quickly, the homophily ratio of the reconstructed graph varies greatly, and the DRL module receives a large reward value in each episode according to Eq.~\ref{equation10}. In the later stage of model training, GNN gives stable classification results, and DRL outputs a stable number of homogeneous adjacent nodes. Since the accuracy and loss of GNN converge to a stable stage, the DRL module converges to the mean reward zero. This suggests that the proposed GraphRARE framework can optimize the graph topology adaptively.

\subsection{Graph Topology Optimization Analysis}\label{sec:topology_optimization}
\begin{figure}[!t]
	\centering 
	\includegraphics[width=\linewidth]{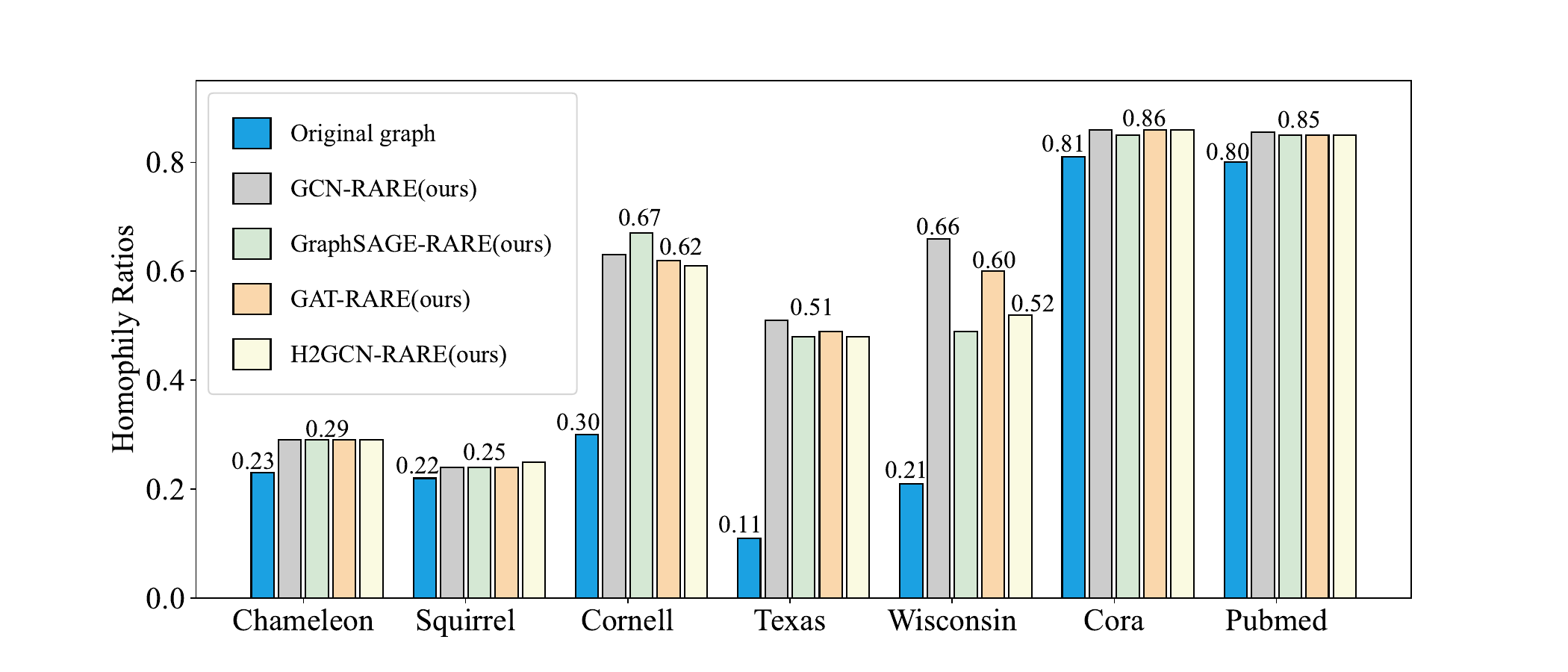} 	
	\caption{Comparisons of the homophily ratios between the original graph and the optimized graphs given by the GraphRARE models on the seven datasets.} 
	\label{fig:homophily}
\end{figure}

The GNNs based on a node message aggregation mechanism are effective under the homophily assumption. Intuitively, the GNN's performance can be improved by increasing the homophily ratio of a graph. The proposed GraphRARE framework can achieve this goal by training the GNNs and DRL jointly based on the relative entropy to optimize the graph topology. The effectiveness of the proposed GraphRARE framework is demonstrated through the homophily ratio comparison between the original and optimized graph topologies. The GCN-RARE, GraphSAGE-RARE, GAT-RARE, and H2GCN-RARE are evaluated on all seven datasets. The results are presented in Figure~\ref{fig:homophily}. 
{
The subdued enhancements in the homophily ratio for the Chameleon and Squirrel datasets could be attributed to their intricate graph topology. Notably, these datasets exhibit denser connectivity compared to the other five datasets, with nodes, on average, having about eight times more connections. This increased density complicates the task of identifying clear patterns and understanding node similarities, thereby posing challenges in achieving notable enhancements in the homophily ratio.} 
The four enhanced GraphRARE models all increase the homophily ratio over the original graph by an average of 0.20 (GCN-RARE), 0.17 (GraphSAGE-RARE and GAT-RARE), and 0.18 (H2GCN-RARE). One can see that the optimized graph topologies of the GCN-RARE, GraphSAGE-RARE, GAT-RARE, and H2GCN-RARE improve the homophily ratio, indicating the effectiveness and generalization ability of the proposed framework.

\subsection{{Relative Entropy Analysis}}
\begin{figure}[ht!]
     \centering
         \begin{subfigure}{0.24\textwidth}
             \centering
             \includegraphics[width=\textwidth]{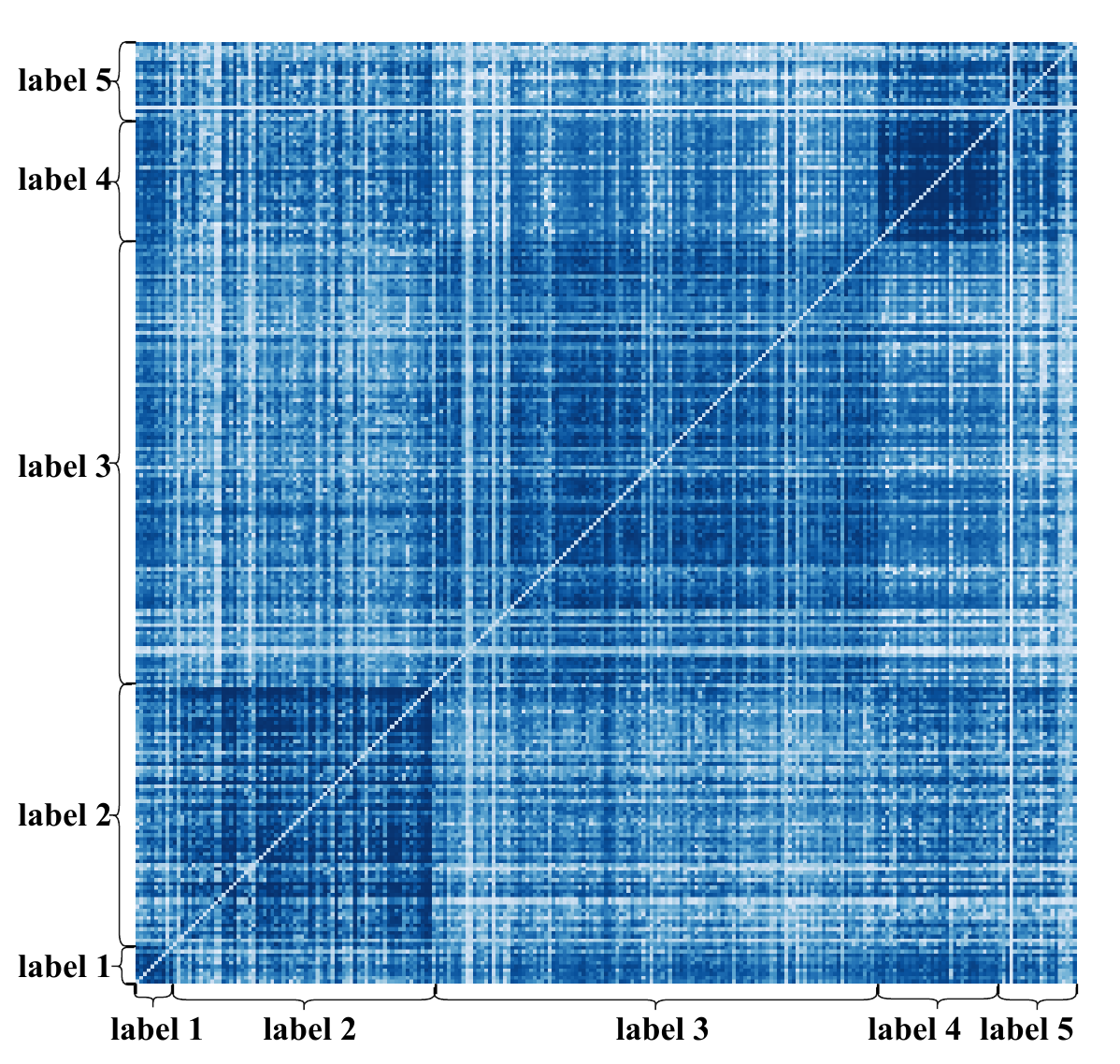}
             \caption{Wisconsin dataset}
         \end{subfigure}
         \begin{subfigure}{0.24\textwidth}
             \centering
             \includegraphics[width=\textwidth]{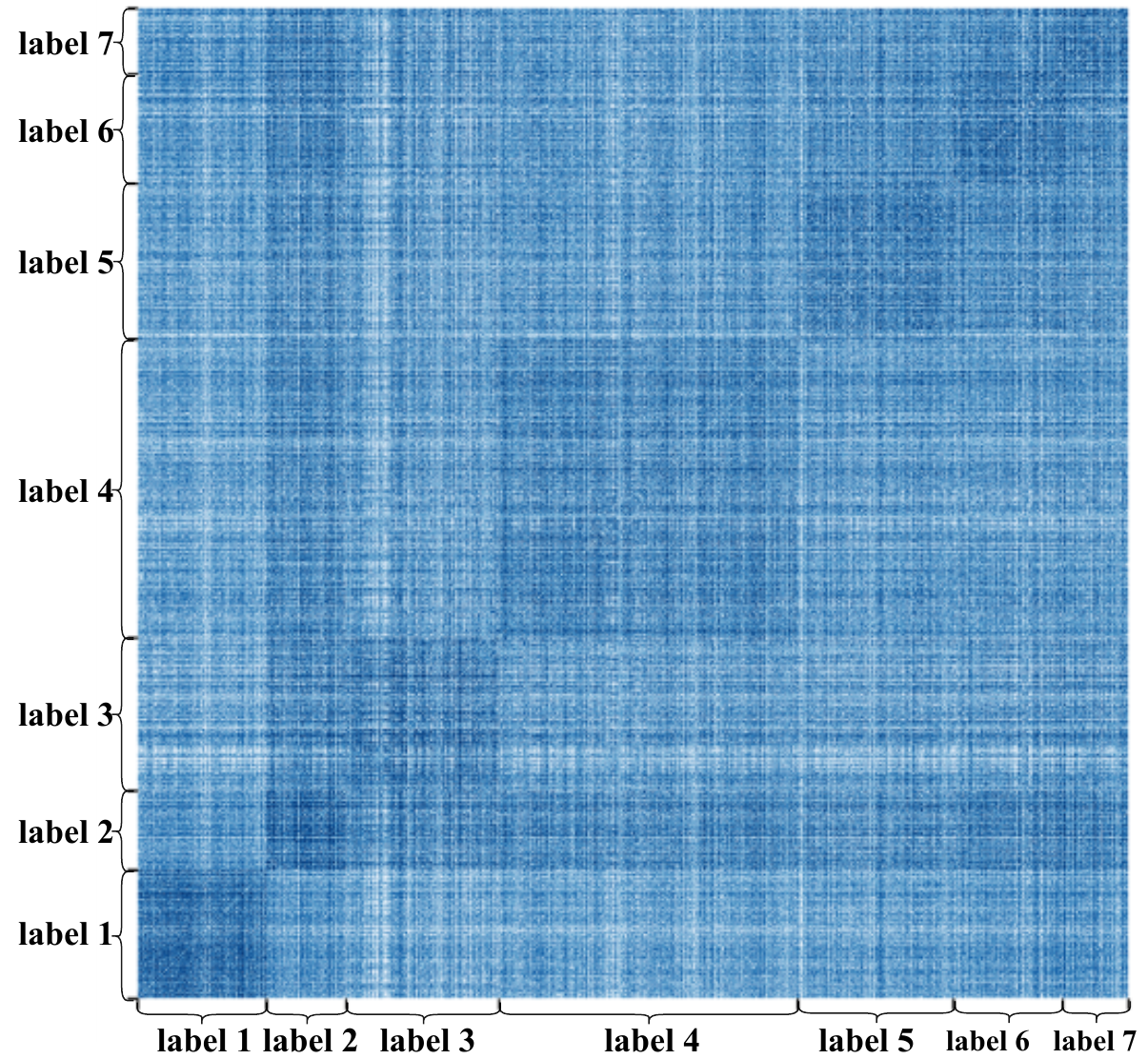}
             \caption{Cora dataset}
        \end{subfigure}
        \caption{{The visualization of node relative entropy on Wisconsin and Cora datasets. Deeper colors indicate higher entropy value between node pairs.}}
        \label{fig:vis_entropy}
\end{figure}

{
Since the node relative entropy computation is an essential module in our framework, we conduct a simple but intuitive visualization experiment to demonstrate its effectiveness. As shown in the Figure~\ref{fig:vis_entropy}, we visualize the node relative entropy between node pairs, where deep colors indicate higher entropy values. Notably, pairs of nodes that share the same label exhibit higher node relative entropy. Consequently, our GraphRARE would connect the these high-entropy node pairs, aligning with the homophily assumption.
}

\section{Related Work}
\noindent\textbf{Graph Neural Networks.} Most existing GNNs adopt the message passing framework and use permutation invariant local aggregation schemes to update node representations. For instance, GCN~\cite{DBLP:conf/iclr/KipfW17} averages features of all neighboring nodes. GAT~\cite{DBLP:conf/iclr/VelickovicCCRLB18} uses an attention mechanism to assign different weights to neighboring nodes. GraphSAGE~\cite{DBLP:conf/nips/HamiltonYL17} samples fixed-size neighbors of a node and aggregates their features for realizing fast and scalable GNN training. However, traditional GNNs suffer from a severe performance reduction under the heterophily setting. 

\noindent\textbf{Relative Entropy on Graph Data.} Relative entropy on graph data is a measure of distance between nodes, which is calculated based on the probability distribution of a node set~\cite{cover1991entropy}. This metric can be used to rank nodes by importance~\cite{DBLP:journals/entropy/OmarP20}. The measurement of relative entropy between nodes has been mainly based on structural parameters, such as the ego node and its neighboring nodes' degrees. Tian et al.~\cite{conf/www/migcn} define the node joint information entropy and node mutual information between a node and its remote neighbors. Zhang et al.~\cite{zhang2018measure} measure the similarity between nodes based on the relative entropy of node degree. In addition to structural information of nodes, node features are also important metrics in measuring node information as they contain rich semantic information (e.g., the research titles in citation networks). However, fewer studies~\cite{luo2021graph, DBLP:conf/aaai/WangJWHH22} have considered node features and a local topological structure jointly to measure mutual information between node pairs. Considering node features and structural information, the proposed framework introduces a node relative entropy to measure the similarity between node pairs.

\noindent\textbf{Graph Topology Optimization.} In recent years, research efforts have been conducted on GNNs with heterophily. Their main idea is to optimize the original graph topology by extending local neighbors to higher-order neighbors. UGCN~\cite{DBLP:conf/nips/JinYHWWHH21} connects the $top\text{-}k$ similar node pairs to reconstruct the graph topology by measuring the feature-level cosine similarity between the ego node and remote nodes using the kNN algorithm. In addition, through the node similarity analysis, SimP-GCN~\cite{DBLP:conf/wsdm/JinDW0LT21} selects the $top\text{-}k$ similar node pairs in terms of feature similarity to construct new neighboring node sets. Based on the structural information, MI-GCN~\cite{conf/www/migcn} reconstructs the graph topology by calculating mutual information between node pairs and setting a fixed number of $top\text{-}k$ new neighbors and $top\text{-}d$ deleted neighbors. However, the $top\text{-}k$ selection methods have mostly been based on a heuristic that the value of $k$ is diverse in different graphs and has to be chosen adaptively, which will cost many human efforts and can result in a mixture of useful and irrelevant information of the multi-hop nodes. Additionally, it is crucial to address the presence of noisy edges in the original graph as they can impact the performance of GNNs. The original graph topology optimization methods strongly rely on prior knowledge or require conducting massive experiments to tune the $k$ value. However, these methods fail to achieve optimal results through end-to-end training. To address this shortcoming, the proposed framework adopts a deep reinforcement learning-based algorithm to choose a proper value of $k$ and $d$ for each node in a graph, which can be trained jointly with the GNN in an end-to-end manner.

{
Deep neural networks is typically modeled as computational graphs, and graph optimization techniques offer structured and efficient computational graph representations that enhance reasoning and elevate training efficiency within the expansive field of database technology. MetaFlow~\cite{DBLP:conf/mlsys/JiaTWGZA19} utilizes a rule-based approach to uncover additional optimization opportunities in computation graphs. Tensor Comprehensions~\cite{DBLP:journals/corr/abs-1802-04730} uses black-box auto-tuning and polyhedral optimizations. Additionally, Fang et al.~\cite{DBLP:journals/pvldb/FangSW020} propose a pruning algorithm and a dynamic programming approach to refine computation graph optimization further. Thus, the optimized graph topology not only enhances the performance of GNN models on downstream tasks but also contributes to the broader field of database technology by offering a more organized and efficient data representation. }

\section{Conclusions}
This work focuses on heterophilic graphs and designs an innovative framework named by GraphRARE. In heterophilic graphs, linked nodes have different features and class labels, whereas the semantically related nodes might be multi-hop away, which can lead to poor GNN's performance. The main objective of the GraphRARE framework is to discover remote but informative nodes based on the defined node relative entropy. In the framework, the DRL module and GNN are combined to optimize the graph topology in an end-to-end manner. Experimental results demonstrate that the proposed framework can be easily adapted to the existing GNN models and can improve their performance on the benchmarks. In future works, a number of extensions and potential improvements are possible, such as extending GraphRARE to incorporate multi-modal graphs or spatial-temporal graphs.

\section{Acknowledgments}
This work is supported in part by the National Key R\&D Program of China (Grant No.2021ZD0112901), National Natural Science Foundation of China (Grant No.6227073648) and the Beijing Postdoctoral Research Foundation (No.2023-22-97). Zhifeng Bao is supported in part by ARC DP240101211 and DP220101434.
\bibliographystyle{IEEEtranS}
\balance
\bibliography{main}

\end{document}